%% file: main.tex
\documentclass[sigconf]{acmart}

\usepackage{subfigure}
\usepackage{multirow} 
\usepackage{amsthm}
\usepackage[ruled, vlined]{algorithm2e}
\usepackage{subfiles}
\usepackage{hyperref}
\usepackage{setspace}
\usepackage{amsmath}

\graphicspath{{figures/}{../figures/}}

\makeatletter
\g@addto@macro\normalsize{%
	\setlength\abovedisplayskip{2pt}
	\setlength\belowdisplayskip{2pt}
	\setlength\abovedisplayshortskip{1pt}
	\setlength\belowdisplayshortskip{1pt}
}
\makeatother

\usepackage{enumitem}
\setenumerate[1]{itemsep=0pt,partopsep=0pt,parsep=\parskip,topsep=1pt}
\setitemize[1]{itemsep=0pt,partopsep=0pt,parsep=\parskip,topsep=1pt}
\setdescription{itemsep=0pt,partopsep=0pt,parsep=\parskip,topsep=1pt}

\newcommand{\mc}[1]{\textcolor{blue}{#1}}
\newcommand{\todo}[1]{\textcolor{red}{[#1]}}
\newcommand{\nop}[1]{}

\newtheorem{definition}{Definition}

\newtheorem{theorem}{Theorem}

\newtheorem{claim}{Claim}
\newtheorem{assumption}{Assumption}

\copyrightyear{2020}
\acmYear{2020}
\setcopyright{acmcopyright}\acmConference[KDD '20]{Proceedings of the 26th ACM SIGKDD Conference on Knowledge Discovery and Data Mining USB Stick}{August 23--27, 2020}{Virtual Event, USA}
\acmBooktitle{Proceedings of the 26th ACM SIGKDD Conference on Knowledge Discovery and Data Mining USB Stick (KDD '20), August 23--27, 2020, Virtual Event, USA}
\acmPrice{15.00}
\acmDOI{10.1145/3394486.3403203}
\acmISBN{978-1-4503-7998-4/20/08}

\AtBeginDocument{%
	\providecommand\BibTeX{{%
			\normalfont B\kern-0.5em{\scshape i\kern-0.25em b}\kern-0.8em\TeX}}}

\renewcommand{\authors}{}

\begin{document}
\begin{sloppy}
\fancyhead{}

\title{Measuring Model Complexity of Neural Networks with Curve Activation Functions}

\author{Xia Hu$^*$}
\affiliation{
  \institution{Simon Fraser University, Burnaby, Canada}
}
\email{huxiah@sfu.ca}

\author{Weiqing Liu}
\affiliation{
  \institution{Microsoft Research, Beijing, China}
}
\email{Weiqing.Liu@microsoft.com}

\author{Jiang Bian}
\affiliation{
  \institution{Microsoft Research, Beijing, China}
}
\email{Jiang.Bian@microsoft.com}

\author{Jian Pei}
\affiliation{
  \institution{Simon Fraser University, Burnaby, Canada}
}
\email{jpei@cs.sfu.ca}

\thanks{$^*$This work was done during Xia Hu's internship at Microsoft Research. \\
	Xia Hu and Jian Pei's research is supported in part by the NSERC Discovery Grant program. All opinions, findings, conclusions and recommendations in this paper are those of the authors and do not necessarily reflect the views of the funding agencies.}

\renewcommand{\shortauthors}{}

\begin{abstract}	
	It is fundamental to measure model complexity of deep neural networks. \nop{A good model complexity measure can help to tackle many challenging problems, such as overfitting detection, model selection, and performance improvement.}  The existing literature on model complexity mainly focuses on neural networks with piecewise linear activation functions. Model complexity of neural networks with general curve activation functions remains an open problem. To tackle the challenge, in this paper, 
	we first propose \emph{linear approximation neural network} (\emph{LANN} for short), a piecewise linear framework to approximate a given deep model with curve activation function. LANN constructs individual piecewise linear approximation for the activation function of each neuron, and minimizes the number of linear regions to satisfy a required approximation degree. Then, we analyze the upper bound of the number of linear regions formed by LANNs, and derive the complexity measure based on the upper bound. 
	To examine the usefulness of the complexity measure, we experimentally explore the training process of neural networks and detect overfitting. Our results demonstrate that the occurrence of overfitting is positively correlated with the increase of model complexity during training. We find that the $L^1$ and $L^2$ regularizations suppress the increase of model complexity. Finally, we propose two approaches to prevent overfitting by directly constraining model complexity, namely neuron pruning and customized $L^1$ regularization. 
\end{abstract}


\maketitle

\section{Introduction}

\subfile{sections/introduction}

\section{Related Work}
\label{sec:relatedwork}

\subfile{sections/relatedwork}

\section{Problem Formulation}

\subfile{sections/overview}

\section{LANN Architecture}
\label{sec:lann}

\subfile{sections/architecture}

\section{Model Complexity}
\label{sec:complex}

\subfile{sections/complexity}

\section{Insight from Complexity}
\label{sec:insight}
\subfile{sections/application}

\section{Conclusion}
\label{sec:conclusion}

\subfile{sections/conclusion}

\bibliographystyle{ACM-Reference-Format}
\bibliography{sections/references}

\clearpage
\appendix

\subfile{sections/appendix}

\end{sloppy}
\end{document}

%% file: sections/introduction.tex
Deep neural networks have gained great popularity in tackling various real-world applications, such as machine translation~\cite{weng2019unsupervised}, speech recognition~\cite{chiu2018state} and computer vision~\cite{gao2019graph}. 
One major reason behind the great success is that the classification function of a deep neural network can be highly nonlinear and express a highly complicated function~\cite{bengio2011expressive}. 
Consequently, a fundamental question lies in how nonlinear and how complex the function of a deep neural network is. Model complexity measures~\cite{montufar2014number, raghu2017expressive} address this question. The recent progress in model complexity measure directly facilitates the advances of many directions of deep neural networks, such as model architecture design, model selection, performance improvement~\cite{hayou2018selection}, and overfitting detection~\cite{hawkins2004problem}. 

The challenges in measuring model complexity are tackled from different angles. For example, the influences of model structure on complexity have been investigated, including layer width, network depth, and layer type. The power of width is discussed and a single hidden layer network with a finite number of neurons is proved to be an universal approximator~\cite{hornik1989multilayer, barron1993universal}. With the exploration of deep network structures, some recent studies pay attention to the effectiveness of deep architectures in increasing model complexity, known as depth efficiency~\cite{lu2017expressive, bengio2011expressive, cohen2016expressive,eldan2016power}. The bounds of model complexity of some specific model structures are proposed, from sum-product networks~\cite{delalleau2011shallow} to piecewise linear neural networks~\cite{pascanu2013number, montufar2014number}. 

\nop{depth and prove \todo{What does this mean?} the exponential advantage of network depth in increasing model complexity}

Model parameters (e.g., weight, bias of layers) also play important roles in model complexity. For example, $f_1(x)=ax+b\sin(x)$ may be considered more complex than $f_2(x)=cx+d$ according to their function forms. However if the parameters of the two functions are $a=1$, $b=0$, $c=1$, and $d=0$, $f_1$ and $f_2$ are then two coincident lines. This example demonstrates the importance of model parameters on complexity. 
\citet{raghu2017expressive} propose a complexity measure for neural networks with piecewise linear activation functions by measuring the number of linear regions through a trajectory path between two instances. Their proposed complexity measure reflects the effect of model parameters to some degree. 

However, the approach of~\cite{raghu2017expressive} cannot be directly generalized to neural networks with curve activation functions, such as Sigmoid~\cite{kilian1993power}, Tanh~\cite{kalman1992tanh}. At the same time, in some specific applications, curve activation functions are found superior than piecewise linear activation functions. For example, many financial models use Tanh rather than ReLU~\cite{ding2015deep}. A series of state-of-the-art studies speed up and simplify the training of neural networks with curve activation functions~\cite{ioffe2015batch}. This motivates our study on model complexity of deep neural networks with curve activation functions. 

\nop{along with methods such as Batch Norm~\cite{ioffe2015batch} improving the training efficiency and preventing from gradient vanishing of networks with Sigmoid, Tanh, the effectiveness of these activation functions is refocused. On the other hand, the effectiveness of different activation functions is inconclusive~\cite{hayou2018selection}. The selection of activation function in practice is mostly based on the attempt, curve activation functions (e.g. Sigmoid, Tanh) still play important roles in practical applications~\cite{ding2015deep}. }

\begin{figure}[t]
	\begin{center}
		\subfigure[]{
			\includegraphics[width = 0.35\linewidth]{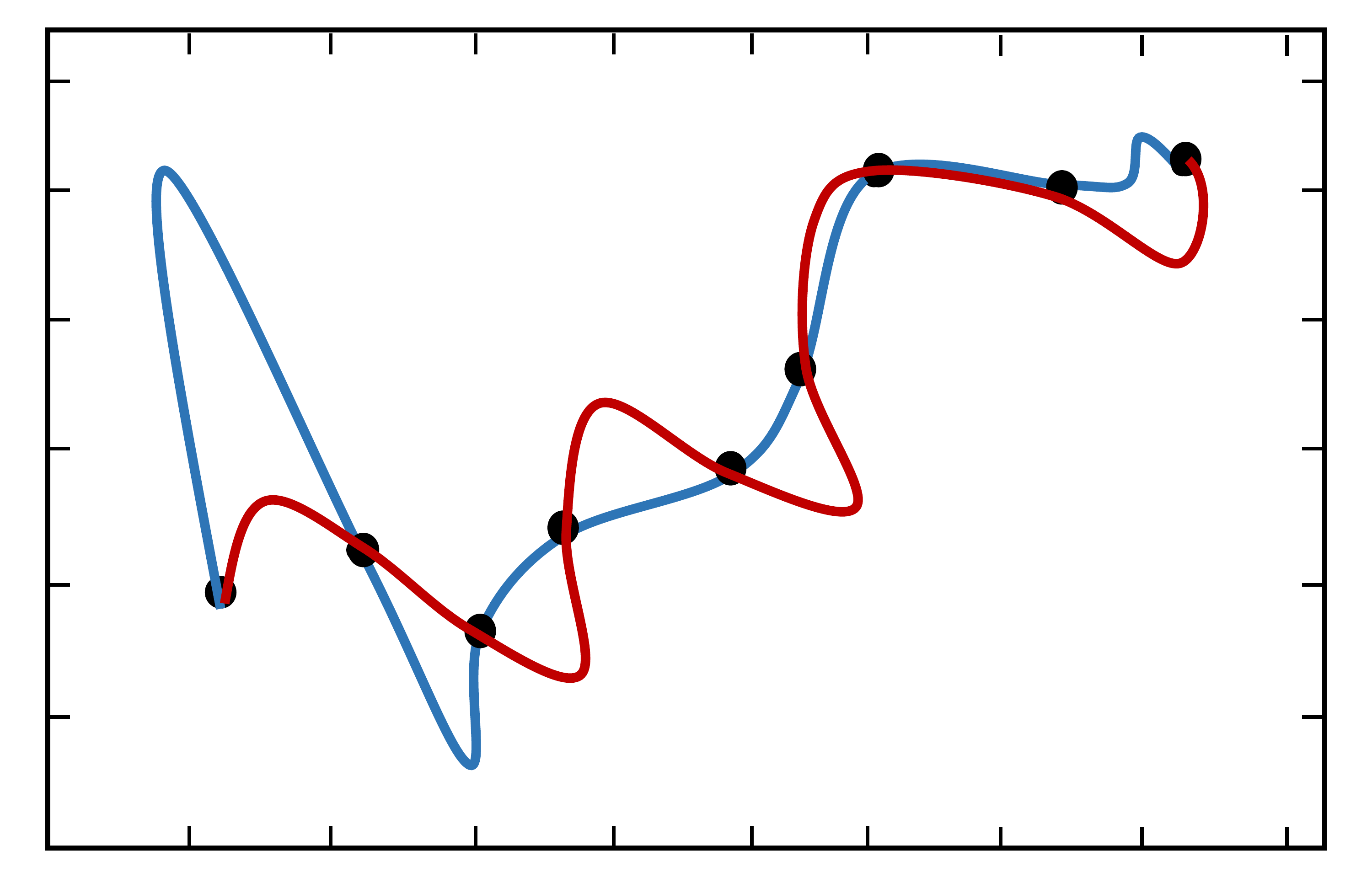}
			\label{fig:intro:mimic}
		}
		\hskip .25in
		\subfigure[]{
			\includegraphics[width = 0.35\linewidth]{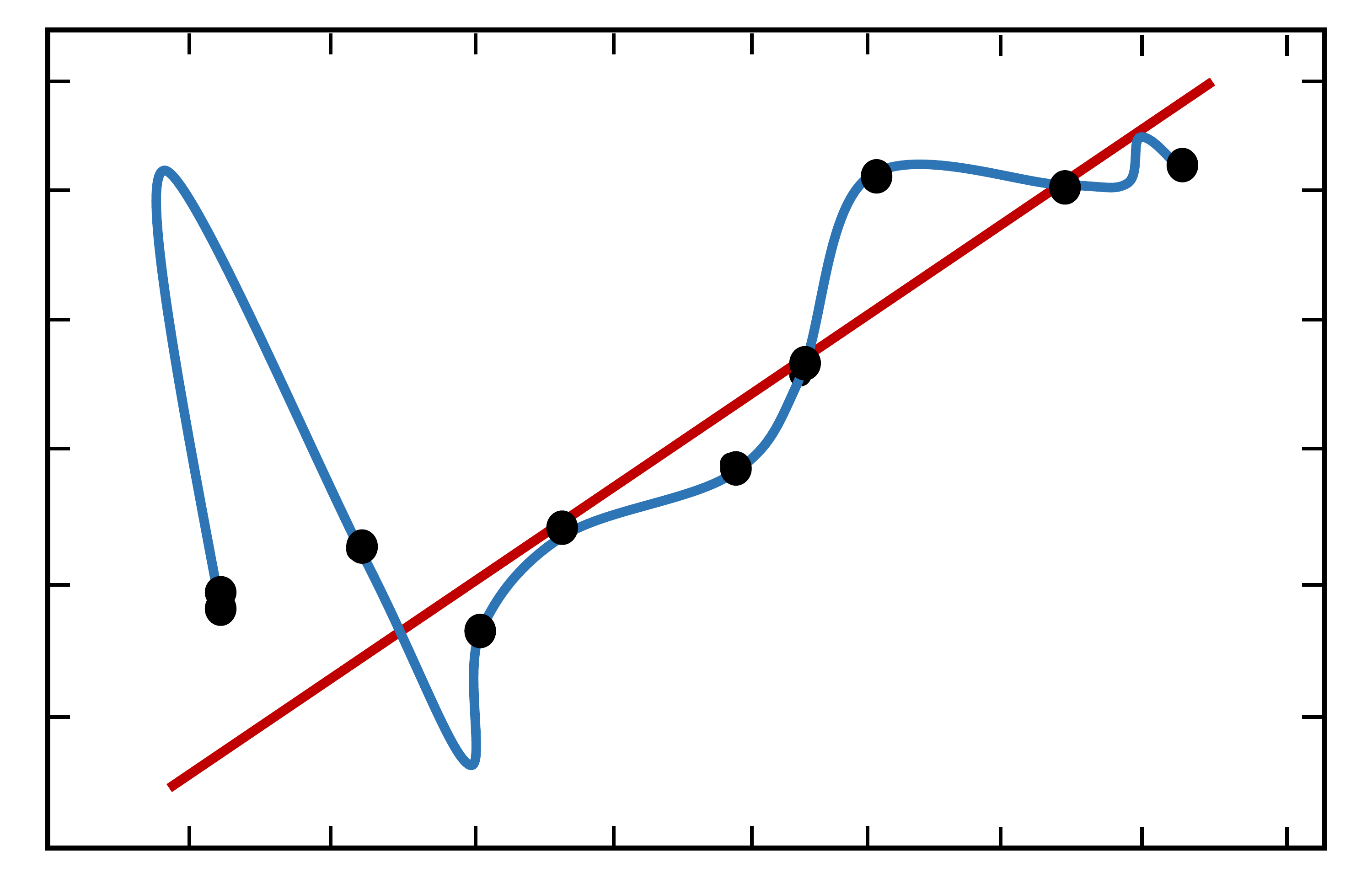}
			\label{fig:intro:overfitting}
		}
		\caption{(a) Two functions behaving similarly on given points may be very different. (b) Illustration of overfitting.  }
		\label{fig:intro}
	\end{center}
\end{figure}

In this paper, we develop a complexity measure for deep fully-connected neural networks with curve activation functions. 
Previous studies on deep models with piecewise linear activation functions use the number of linear regions to model the nonlinearity and measure model complexity~\cite{pascanu2013number,montufar2014number,raghu2017expressive,novak2018sensitivity}. To generalize this idea, we develop a piecewise linear approximation to approach target deep models with curve activation functions. Then, we measure the number of linear regions of the approximation as an indicator of the target model complexity. The piecewise linear approximation is designed under two desiderata. First, to guarantee the approximation degree \nop{with respect to the target model}, we require a direct approximation of the function of the target model rather than simply mimicking the behavior or performance, such as the mimic learning approach~\cite{hinton2015distilling}.  The rationale is that two functions having the same behavior on a set of data points may still be very different, as illustrated in Figure~\ref{fig:intro:mimic}.  Therefore, approximation using the mimic learning approach~\cite{hinton2015distilling} is not enough. Second, to compare the complexity values of different models, the complexity measure has to be principled. The principle we follow is to minimize the number of linear regions given an approximation degree threshold. Under these two desiderata, the minimum number of linear regions constrained by a certain approximation degree can be used to reflect the model complexity. 


Technically we propose the \emph{linear approximation neural network} (\emph{LANN} for short), a piecewise linear framework to approximate a target deep model with curve activation functions. A LANN shares the same layer width, depth and parameters with the target model, except that it replaces every activation function with a piecewise linear approximation. An individual piecewise linear function is designed as the activation function on every neuron to satisfy the above two desiderata. We analyze the approximation degree of LANNs with respect to the target model, then devise an algorithm to build LANNs to minimize the number of linear regions\nop{of approximation}. We provide an upper bound on the number of linear regions formed by LANNs, and define the complexity measure using the upper bound. 

To demonstrate the usefulness of the complexity measure, we explore its utility in analyzing the training process of deep models, especially the problem of overfitting~\cite{hawkins2004problem}.  Overfitting occurs when a model is more complicated than the ultimately optimal one, and thus the learned function fits too closely to the training data and fails to generalize, as illustrated in Figure~\ref{fig:intro:overfitting}.
Our results show that the occurrence of overfitting is positively correlated to the increase of model complexity. Besides, we observe that regularization methods for preventing overfitting, such as $L^1$ and $L^2$ regularizations~\cite{goodfellow2016deep}, constrain the increase of model complexity. Based on this finding, we propose two simple yet effective approaches for preventing overfitting by directly constraining the growth of model complexity.

The rest of the paper is organized as follows.  Section~\ref{sec:relatedwork} reviews related work. In Section 3 we provide the problem formulation. In Section~\ref{sec:lann} we introduce the linear approximation neural network framework. In Section~\ref{sec:complex} we develop the complexity measure. In Section~\ref{sec:insight} we explore the training process and overfitting in the view of complexity measure. Section~\ref{sec:conclusion} concludes the paper.

%% file: sections/relatedwork.tex
The studies of model complexity dates back to several decades. In this section, we review related works of model complexity of neural networks from two aspects: model structures and parameters. 

\subsection{Model Structures}
Model structures may have strong influence on model complexity, such as width, layer depth, and layer type. 

The power of layer width of shallow neural networks is investigated~\cite{hornik1989multilayer, barron1993universal, cybenko1989approximation, maass1994comparison} decades ago. \citet{hornik1989multilayer} propose the universal approximation theorem, which states that a single layer feedforward network with a finite number of neurons can approximate any continuous function under some mild assumptions. Some later studies~\cite{barron1993universal, cybenko1989approximation, maass1994comparison} further strengthen this theorem. However, although with the universal approximation theorem, the layer width can be exponentially large. 
\citet{lu2017expressive} extend the universal approximation theorem to deep networks with bounded layer width. 

Recently, deep models are empirically discovered to be more effective than a shallow one. \nop{Subsequently,}A series of studies focus on exploring the advantages of deep architecture in a theoretical view, which is called depth efficiency~\cite{ bengio2011expressive,cohen2016expressive,poole2016exponential,eldan2016power}.
Those studies show that the complexity of \nop{the function of} a deep network can only be matched by a shallow one with exponentially more nodes. In other words, the function of deep architecture achieves exponential complexity in depth while incurs polynomial complexity in layer width. 

Some studies bound the model complexity with respect to certain structures or activation functions~\cite{delalleau2011shallow, du2018power, montufar2014number, bianchini2014complexity, poole2016exponential}. 
\citet{delalleau2011shallow} study sum-product networks and use the number of monomials to reflect model complexity. \citet{pascanu2013number} and \citet{montufar2014number} investigate fully connected neural networks with piecewise linear activation functions (e.g. ReLU and Maxout), and use the number of linear regions as a representation of complexity. \nop{They theoretically bound the number of linear regions.} However, the studies on model complexity only from structures are not able to distinguish differences between two models with similar structures, which are needed for problems such as understanding model training.

\subsection{Parameters}

Besides structures, the value of model parameters, including layer weight and bias, also play a central role in model complexity measures. Complexity of models is sensitive to the values of parameters. 

\citet{raghu2017expressive} propose a complexity measure for DNNs with piecewise linear activation functions. They follow the previous studies on DNNs with piecewise linear activation functions and use the number of linear regions as a reflection of model complexity~\cite{pascanu2013number, montufar2014number}. To measure how many linear regions a data manifold is split, \citet{raghu2017expressive} build a trajectory path from one input instance to another, then estimate model complexity by the number of linear region transitions through the trajectory path. Their trajectory length measure not only reflects the influences of model structures on model complexity, but also is sensitive to model parameters. They further study Batch Norm~\cite{ioffe2015batch} using the complexity measure. Later, \citet{novak2018sensitivity} generalize the trajectory measure to investigate the relationship between complexity and generalization of DNNs with piecewise linear activation functions. 

However, the complexity measure using trajectory~\cite{raghu2017expressive} cannot be directly generalized to curve activation functions. In this paper, we propose a complexity measure to DNNs with curve activation functions by building its piecewise linear approximation. Our proposed measure can reflect the influences of both model structures and parameters.
\nop{
Following the idea of~\cite{pascanu2013number, montufar2014number, raghu2017expressive, novak2018sensitivity}, we use the number of linear regions as a representation of model complexity. There are two reasons. 
First, since the number of linear regions reflects the nonlinearity of a piecewise linear model~\cite{montufar2014number, raghu2017expressive}, the number of linear regions used to approximate a curve function can be considered a reflection of the complexity of the curve function. 
Second, linear regions are finite numbers and are detectable with the status of piecewise linear activation functions. \nop{This makes it a useful metric.}
}

%% file: sections/overview.tex
A \textbf{deep (fully connected) neural network} (\textbf{DNN} for short) consists of a series of fully connected layers. Each layer includes an affine transformation and a nonlinear activation function. 
In classification tasks, let $f:\mathbb{R}^d \rightarrow \mathbb{R}^c$ represent a DNN model, where $d$ is the number of features of inputs, and $c$ the number of class labels. For an input instance $x \in \mathbb{R}^d$, $f$ can be written in the form of
\begin{equation}
	f(x)  = V_oh_{L}(h_{L-1}(\cdots(h_1(x))))+b_o
\end{equation}
where $V_o$ and $b_o$, respectively, are the weight matrix and the bias vector of the output layer, $f(x) \in \mathbb{R}^c$ is the output vector corresponding to the $c$ class labels, $L$ is the number of hidden layers, and $h_i$ is $i$-th hidden layer in the  form of
\begin{equation}
	h_i(z) = \phi (V_iz + b_i),\ \ i=1,\ldots, L
\end{equation}
where $V_i$ and $b_i$ are the weight matrix and the bias vector of the $i$-th hidden layer, respectively. $\phi(\cdot)$ is the activation function. In this paper, if $z$ is a vector, we use $\phi(z)$ to represent the vector obtained by separately applying $\phi$ to each element of $z$. 
\nop{$m_i$ denotes the number of neurons of $i$-th hidden layer. The training set is defined as $\mathcal{D} = \{(x, y)\}^N$ where $x \in \mathcal{X}$ is a vector of features and $y \in \{0, \ldots, c-1\}$ is the class label. }
\nop{Let $f_{\sigma} = \sigma(f(x))$ denote the prediction probability of forwarding $f(x)$ to softmax function $\sigma(\cdot)$.}

The commonly used activation functions can be divided into two groups according to algebraic properties. First, a \textbf{piecewise linear activation function} is composed of a finite number of pieces of affine functions. Some commonly used piecewise linear activation functions include ReLU~\cite{nair2010rectified} and hard Tanh~\cite{nwankpa2018activation}. With a piecewise linear $\phi$, the DNN model $f$ is a continuous piecewise linear function.  Second, a \textbf{curve activation function} is a continuous nonlinear function whose geometric shape is a smooth curved line. Commonly used curve activation functions include Sigmoid~\cite{kilian1993power} and Tanh~\cite{kalman1992tanh}. With a curvilinear $\phi$, the DNN model $f$ is a curve function.  

In this paper, we are interested in fully connected neural networks with curve activation functions. We focus on two typical curve activation functions, Sigmoid~\cite{kilian1993power}, Tanh~\cite{kalman1992tanh}. Our methodology can be easily extended to other curve activation functions.

\begin{figure}[t]
	\begin{center}
		\subfigure[Approximation $g_1$]{
			\includegraphics[width = 0.4\linewidth]{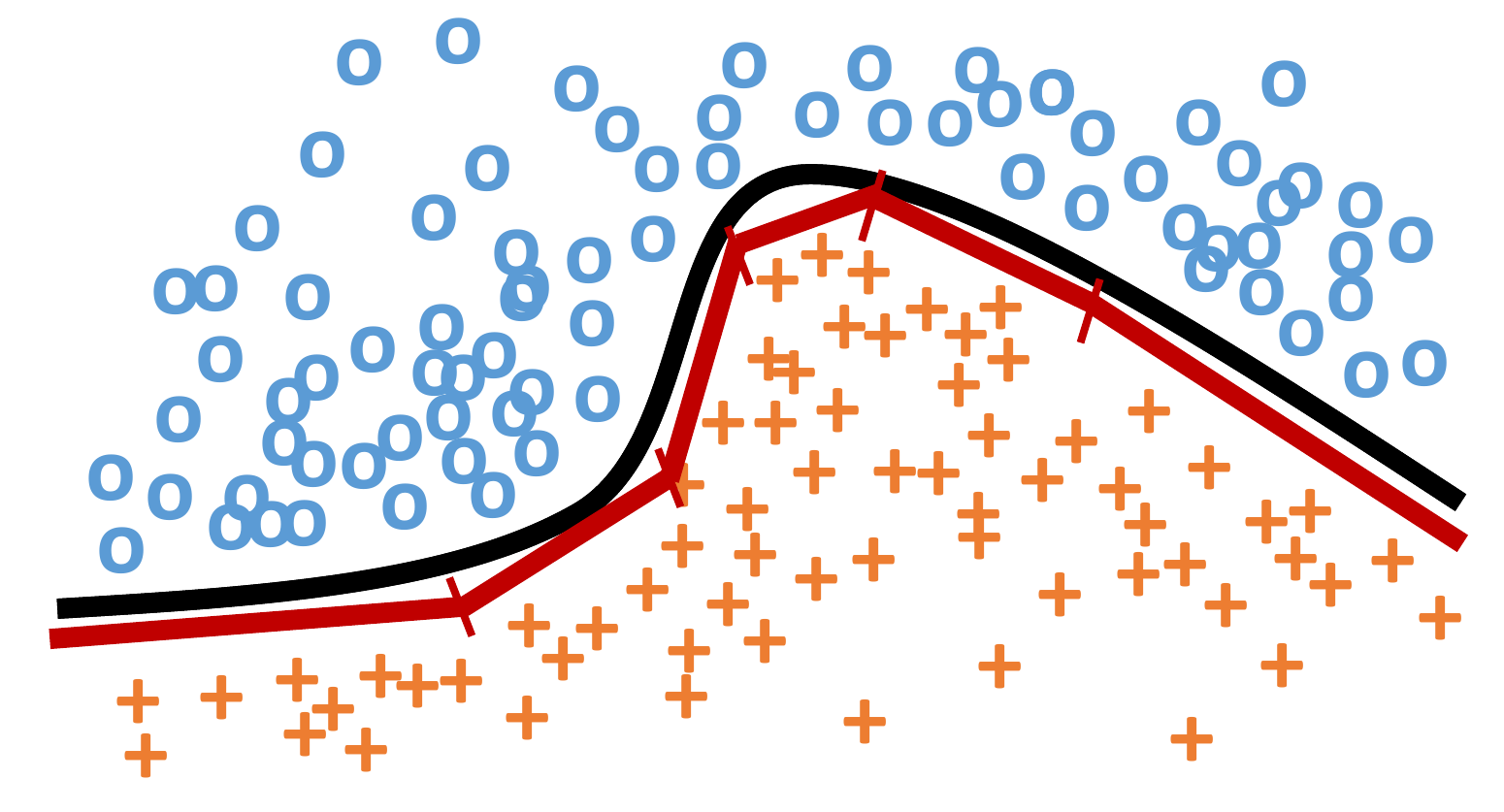}
			\label{fig:universalprinciple:1}
		}
		\hspace{.2in}
		\subfigure[Approximation $g_2$]{
			\includegraphics[width=0.4\linewidth]{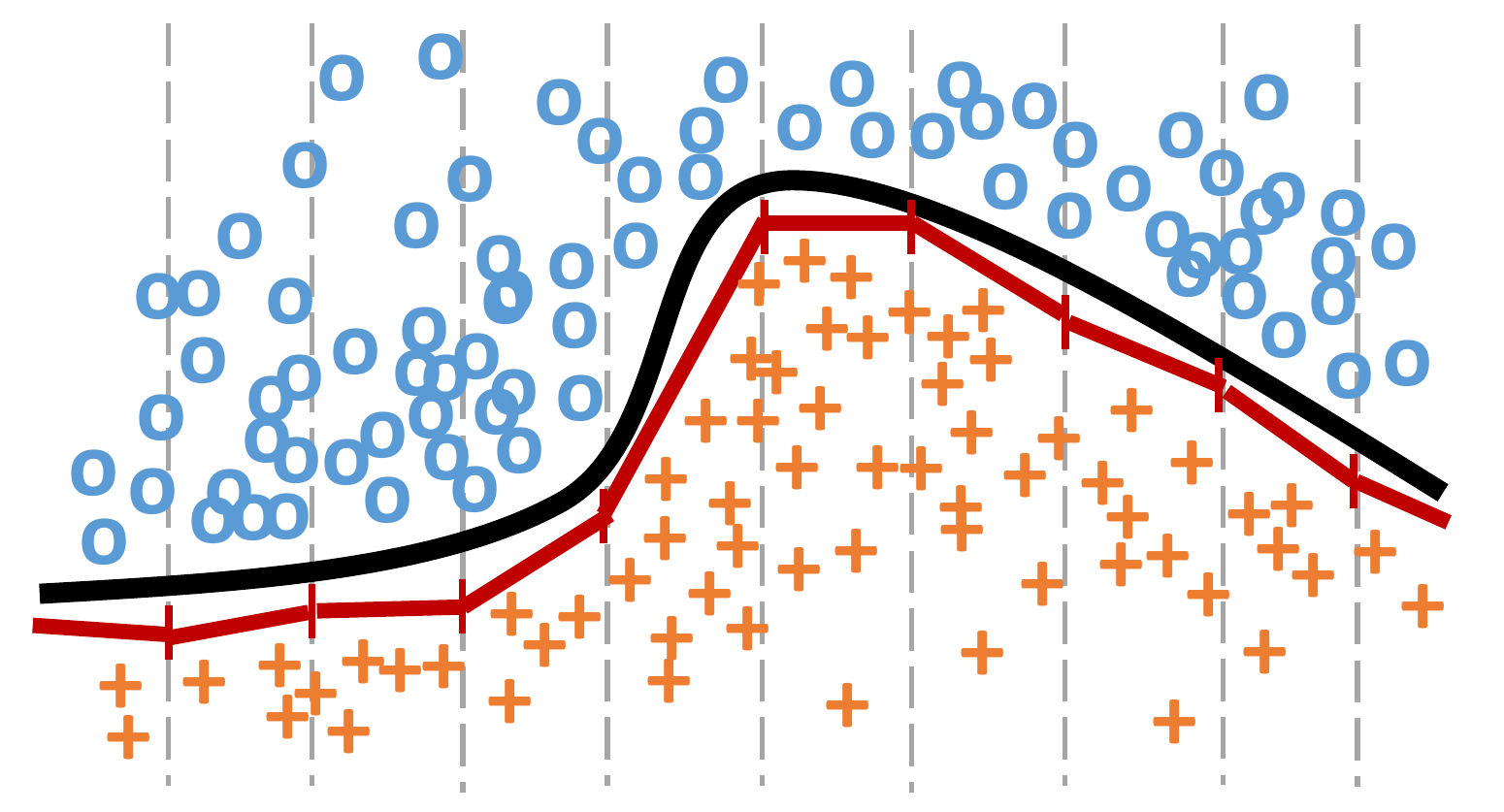}
			\label{fig:universalprinciple:2}
		}
		\caption{Example shows piecewise linear approximation under different approximation principles. }
		\label{fig:universalprinciple}
	\end{center}
\end{figure}

Given a \textbf{target model}, which is a trained fully connected neural network with curve activation functions, we want to measure the model complexity. Here, the complexity reflects how nonlinear, or how curved the function of the network achieves. Our complexity measure should take both the model structure and the parameters into consideration. 
To measure the model complexity, our main idea is to obtain a piecewise linear approximation of the target model, then use the number of linear segments of approximation to reflect the target model complexity. This idea is inspired by the previous studies on DNNs with piecewise linear activation functions~\cite{montufar2014number,novak2018sensitivity,raghu2017expressive}. 
To make our idea of measuring by approximation feasible, the approximation should satisfy two requirements.

First, \emph{the quality/degree of approximation should be guaranteed.} To make the idea of measuring complexity by the nonlinearity of approximation feasible, a prerequisite is that the approximation should be highly close to the function of the target model. In this case, the mimic learning approach~\cite{hinton2015distilling}, which approximates by learning a student model under the guidance of the target model outputs, is not suitable, since it learns the behavior of the target model on a specific dataset and cannot guarantee the generalizability, as illustrated in Figure~\ref{fig:intro:mimic}.  To ensure the closeness of the approximation functions to the target models, we propose \emph{linear approximation neural network} (LANN). A LANN is an approximation model \nop{to the target model}that builds piecewise linear approximations to activation functions in the target model. To make the approximation degree controllable and flexible, we design an individual approximation function for the activation function on every neuron separately according to their status distributions (Section~\ref{sec:lapprox}). Furthermore, we define a measure of approximation degree in terms of approximation error and analyze through error propagation (Section~\ref{sec:approxerror}).

Second, \emph{the approximation should be constructed in a principled manner.} To understand the rationale of this requirement, consider an example in Figure~\ref{fig:universalprinciple}, where the target model is a curved line~(the solid curve). One approximation $g_1$ (the red line in Figure~\ref{fig:universalprinciple:1}) is built using as few linear segments as possible. Another approximation $g_2$  (the red line in Figure~\ref{fig:universalprinciple:2}) evenly divides the input domain into small pieces and then approximates each piece using linear segments. Both of them can approximate the target model to a required approximation degree and can reflect the complexity of the target model. 
However, we should not use $g_1$ on some occasions and use $g_2$ on some other occasions to measure the complexity of the target model, since they are built following different protocols.  To make the complexity measure comparable, the approximation should be constructed under a consistent protocol. We suggest constructing approximations under the protocol of using as few linear segments as possible (Section~\ref{sec:approxalgo}), an thus the minimum number of linear segments required to satisfy the approximation degree can reflect the model complexity.

%% file: sections/architecture.tex
To develop our complexity measure, we propose LANN, a piecewise linear approximation to the target model. 
In this section, we first introduce the architecture of LANN. Then, we discuss the degree of approximation. Last, we propose the algorithm of building a LANN.

\nop{An important prerequisite of our work is the independent and identically distributed assumption:
	\begin{assumption}[i.i.d.]
		\label{ass:iid}
		The i.i.d. assumption indicates that the training and test data are independent, identically distributed. 
\end{assumption}}

\subsection{Linear Approximation Neural Network} 
\label{sec:lapprox}

The function of a deep model with piecewise linear activation functions is piecewise linear, and has a finite number of linear regions~\cite{montufar2014number}. The number of linear regions of such a model is commonly used to assess the nonlinearity of the model, i.e., the complexity~\cite{montufar2014number, raghu2017expressive}.  
Motivated by this, we develop a piecewise linear approximation of the target model with curve activation functions, then use the number of linear regions of the approximation model as a reflection of the complexity of the target model. 


The approximation model we propose is called the \emph{linear approximation neural network} (LANN). 

\begin{definition}[Linear Approximation Neural Network]
	\label{def:lapprox}
	Given a fully connected neural network $f:\mathbb{R}^d \rightarrow \mathbb{R}^c$, a linear approximation neural network $g: \mathbb{R}^d \rightarrow \mathbb{R}^c$ is an approximation of $f$ in which each activation function $\phi(\cdot)$ in $f$ is replaced by a piecewise linear approximation function $\ell(\cdot)$. 
\end{definition}

\nop{\begin{figure}[t]
		\setlength{\abovecaptionskip}{-3pt} 
		\setlength{\belowcaptionskip}{0pt} 
		\begin{center}
			\includegraphics[width=\linewidth]{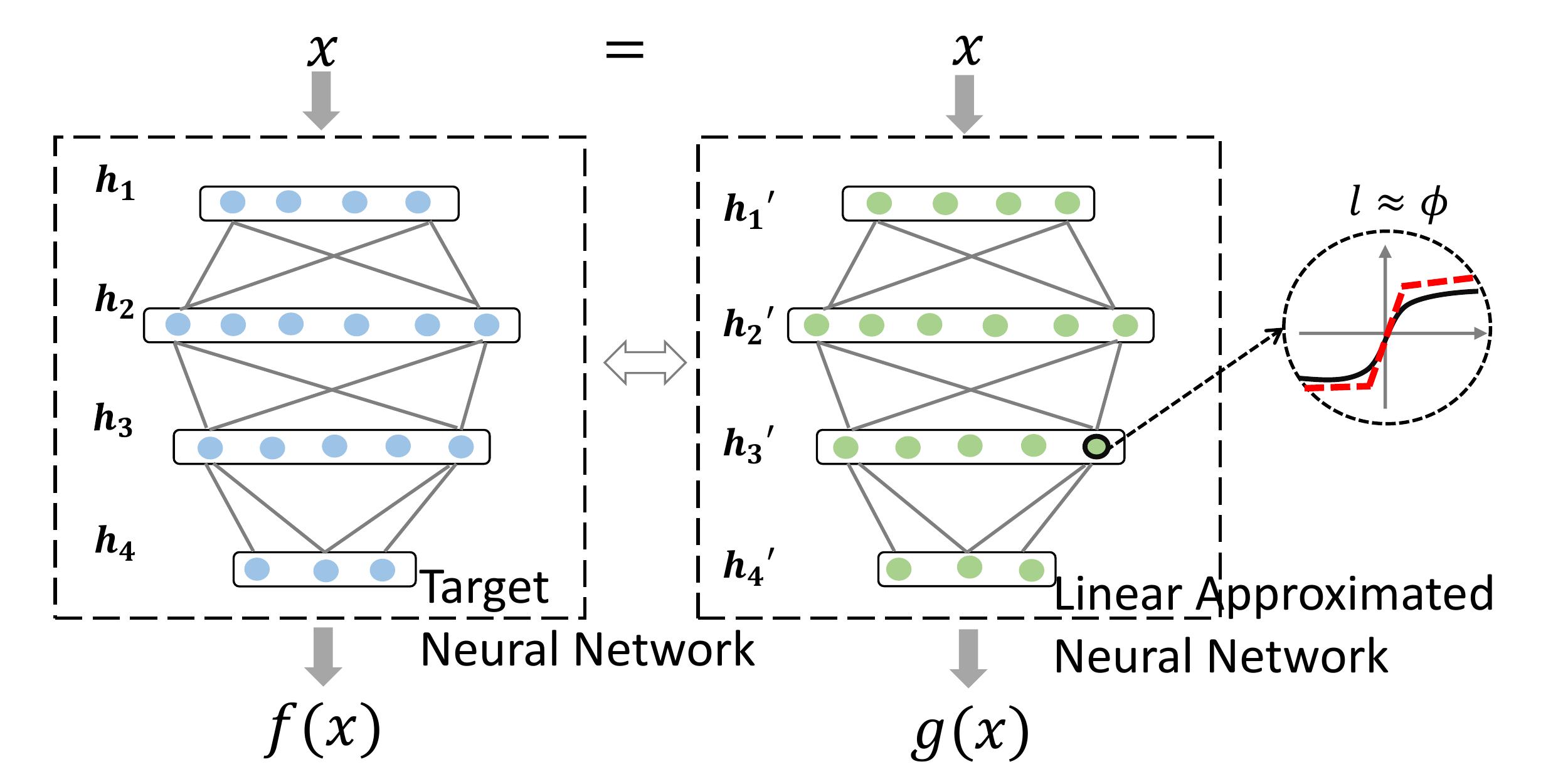}
			\caption{The structure of a LANN. }
			\label{fig:lapproxstructure}
		\end{center}
		\vskip -0.25in
\end{figure}}

A LANN shares the same layer depth, width as well as weight matrix and bias vector as the target model, except that it approximates every activation function using an individual piecewise linear function. This brings two advantages. First, designing an individual approximation function for each neuron makes the approximation degree of a LANN $g$ to the target model $f$ flexible and controllable. Second, the number of subfunctions of neurons is able to reflect the nonlinearity of the network. These two advantages will be further discussed in Section~\ref{sec:approxerror} and Section~\ref{sec:complex}, respectively. 

A piecewise linear function $\ell(\cdot)$ consisting of $k$ subfunctions (linear regions) can be written in the following form.
\begin{equation} 
	\label{eq: linear}
	\ell(z) = \left \{ \begin{array}{rcl}
	\alpha_1z+\beta_1, & \mbox{if} & \eta_0 < z \leq \eta_1 \\
	\alpha_2z+\beta_2, & \mbox{if} & \eta_1 < z \leq \eta_2 \\
	\vdots \\
	\alpha_kz+\beta_k, & \mbox{if} & \eta_{k-1} < z \leq \eta_k
	\end{array} \right.
\end{equation}
where $\alpha_i, \beta_i \in \mathbb{R}$ are the parameters of the $i$-th subfunction. 
Given a variable $z$, the $i$-th subfunction is \textbf{activated} if $z \in (\eta_{i-1}, \eta_i]$, denote by $s(z) = i$. 
Let $\alpha^*=\alpha_{s(z)}$ and $\beta^*=\beta_{s(z)}$ be the parameters of the activated subfunction.  We have $\ell(z) = \alpha^*z+\beta^*$. 

Let $\phi_{i,j}$ be the activation function of the neuron $\{i,j\}$, which represents the $j$-th neuron in $i$-th layer. Then, $\ell_{i,j}$ is the approximation of $\phi_{i,j}$. Let $\ell_i=\{\ell_{i,1}, \ell_{i,2}, \ldots, \ell_{i,m_i}\}$ be the set of approximation functions for $i$-th hidden layer, $m_i$ is the width of $i$-th hidden layer. 
The $i$-th layer of a LANN can be written as
\begin{equation}
	h'_i(z)=\ell_i (V_iz+b_i)
\end{equation}
Then, a LANN is in the form of
\begin{equation}
	g(x) = V_oh'_L(h'_{L-1}(\ldots(h'_1(x)))) + b_o
\end{equation}

Since the composition of piecewise linear functions is piecewise linear, a LANN is a piecewise linear neural network. 
A linear region of the piecewise linear neural network can be represented by the \textbf{activation pattern} (this term follows the convention in ~\cite{raghu2017expressive}):

\begin{definition}[Activation pattern]
	An activation pattern of a piecewise linear neural network is the set of activation statuses of all neurons, denoted by 
	\begin{math}
	s=\{s_{1, 1}, \ldots, s_{1, m_1}, \ldots, s_{L, 1}, \ldots, s_{L, m_L}\},   
	\end{math}
	where $s_{i,j}$ is the activation status of neuron $\{i,j\}$.
\end{definition}

Given an arbitrary input $x$, the corresponding activation pattern $s(x)$ is determined. With the fixed $s(x)$, the transformation of $\ell_i$ of any layer $i$ is reduced to a linear transformation that can be written in the following square matrix.
\begin{equation}
	\small
	L_i = \left[ \begin{array}{ccccc}
	\alpha^*_{i, 1} & 0 &  \ldots & 0 & \beta^*_{i, 1} \\
	0 & \alpha^*_{i, 2} & \ldots & 0 & \beta^*_{i, 2} \\
	\vdots & \vdots &  \ddots & \vdots & \vdots \\
	0  & 0 & \ldots & \alpha^*_{i,m_i} & \beta^*_{i, m_i} \\ 
	0 & 0 & \ldots & 0 & 1
	\end{array}\right ]
\end{equation}
where $\alpha^*_{i,j}$ and $\beta^*_{i,j}$ are the parameters of the activated subfunction of neuron $\{i,j\}$, and are determined by $s_{i,j}$.
The piecewise linear neural network is reduced to a linear function $y=Wx+b$ with
\begin{equation}
	\left[\! \begin{array}{cc} W\! & b \end{array}\!\right] = 
	\left[\! \begin{array}{cc} V_o\! & b_o \end{array}\! \right] 
	\prod_{i = L,\ldots, 1}\!\! \left(L_i
	\left[\! \begin{array}{cc} V_i\! & b_i \\ \textbf{0} & 1\end{array}\! \right] \right)
\end{equation}

An activation pattern corresponds to a linear region of the piecewise linear neural network. Given two different activation patterns, the square matrix $L_i$ of at least one layer are different, so are the corresponding linear functions. Thus, a linear region of the piecewise linear neural network can be expressed by an unique activation pattern. That is, the activation pattern $s(x)$ represents the linear region including $x$.

\subsection{Degree of Approximation} 
\label{sec:approxerror}


We measure the complexity of models with respect to approximation degree.  We first define a measure of approximation degree using approximation error. Then, we analyze approximation error of LANN in terms of neuronal approximation functions. 


\begin{definition}[Approximation error]
	\label{def:approxerror}
	Let $f': \mathbb{R} \rightarrow \mathbb{R}$ be an approximation function of $f: \mathbb{R} \rightarrow \mathbb{R}$. Given input $x$, the approximation error of $f'$ at $x$ is $e(x) = |f'(x) - f(x)|$.
	\label{def:approxerror:g}
	
	Given a deep neural network $f:\mathbb{R}^d \rightarrow \mathbb{R}^c$ and a linear approximation neural network $g: \mathbb{R}^d \rightarrow \mathbb{R}^c$ learned from $f$. We define the approximation error of $g$ to $f$ as the expectation of the absolute distance between their outputs:
	\begin{equation}
	\mathcal{E}(g; f) = \mathbb{E}[\ \frac{1}{c}\sum |g(x) - f(x)|\ ]
	\end{equation}
\end{definition}

A LANN is learned by conducting piecewise linear approximation to every activation function. The approximation of every activation may produce an approximation error. The approximation error of a LANN is the accumulation of all neurons' approximation errors.

In literature~\cite{raghu2017expressive,eldan2016power}, approximation error of activation is treated as a small perturbation added to a neuron, and is observed to grow exponentially through forward propagation. 
Based on this, we go a step further to estimate the contribution of perturbation of every neuron to the model output by analyzing error propagation. 

Consider a target model $f$ and its LANN approximation $g$. According to Definition~\ref{def:approxerror}, the approximation error of $\ell_{i,j}$ of $g$ corresponding to neuron $\{i,j\}$ can be rewritten as $e_{i, j} = |\ell_{i,j} - \phi_{i,j}|$.

Suppose the same input instance is fed into $f$ and $g$ simultaneously. After the forward computation of the first $i$ hidden layers, 
let $r_i$ be the output difference of the $i$-th hidden layer between $g$ and $f$, and $r_{i-1}$ for the $(i-1)$-th layer. Let $x$ denote the input to the $i$-th layer, also the output of the $(i-1)$-th layer of $f$. We can compute $r_i$ by
\begin{equation}
	\label{eq:approxerr:diff}
	r_i = h'_i(x+r_{i-1}) - h_i(x)
\end{equation}
The absolute value of $r_i$ is
\begin{equation}
	\label{eq:approxerror:r}
	\begin{split}
	|r_i|\!    &= |h'_i(x+r_{i-1}) - h_i(x)| \\
	&= |h'_i(x+r_{i-1}) - h_i(x+r_{i-1}) + h_i(x+r_{i-1}) - h_i(x)| \\
	&\leq |h'_i(x+r_{i-1}) - h_i(x+r_{i-1})| +|h_i(x+r_{i-1}) - h_i(x)|
	\end{split}
\end{equation}
To keep the discussion simple, we write $x_r=x+r_{i-1}$. The first term of the righthand side of Eq.~(\ref{eq:approxerror:r}) is
\begin{equation}
	\label{eq:approxerror:rplus1}
	|h'_i(x_r) - h_i(x_r)| = e_i(V_ix_r+b_i),
\end{equation}
where $e_i = [e_{i,1}, e_{i,2}, \ldots, e_{i, m_i}]^T$ is a vector consisting of every neuron's approximation error of the $i$-th layer.
Applying the first-order Taylor expansion to the second term of Eq.~(\ref{eq:approxerror:r}), we have:
\begin{equation}
	\label{eq:approxerror:rplus2}
	|h_i(x+r_{i-1}) - h_i(x)| = |J_i(x)r_{i-1} + \epsilon_i|
\end{equation}
where $J_i(x)=\frac{dh_i(x)}{dx}$ is the Jacobian matrix of the $i$-th hidden layer of $f$, $\epsilon_i$ is the remainder of the first-order Taylor approximation \nop{with the form of $\epsilon_i = H_i(\xi)r_{i-1}^2/2$, $\xi \in [x, x+r_{i-1}]$, $H_i$ is the Hessian matrix of the $i$-the hidden layer of $f$}. Plugging Eq.~(\ref{eq:approxerror:rplus1}) and Eq.~(\ref{eq:approxerror:rplus2}) into Eq.~(\ref{eq:approxerror:r}), we have:
\begin{equation}
	\label{eq:approxerror:rplus}
	|r_i| \leq e_i(V_ix_r+b_i) + |J_i(x)r_{i-1}| + |\epsilon_i|
\end{equation}
Assuming $x$ and $r_{i-1}$ being independent, the expectation of $|r_i|$ is
\begin{equation}
	\mathbb{E}[|r_i|] \leq \mathbb{E}[e_i] + \mathbb{E}[|J_i|]\ \mathbb{E}[|r_{i-1}|] + \mathbb{E}[\hat{\epsilon_i}]
\end{equation}
where the error $\hat{\epsilon_i} = \epsilon_i + \varepsilon_i$, where $\varepsilon_i$ denotes the error in $\mathbb{E}[e_i]$, in other words, the disturbances of $r_{i-1}$ on the distribution of $e_i$. Since $\mathbb{E}[e_i]$ is a vector where the elements correspond to the neurons in the $i$-layer layer, the expectation of $e_{i,j}$ is
\begin{equation}
	\mathbb{E}[e_{i, j}] = \int e_{i, j}(x)t_{i, j}(x)dx,
\end{equation}
where $t_{i, j}(x)$ is \nop{the} probability density function (PDF) of neuron $\{i,j\}$. 

We notice that $h_i(x)$ consists of a linear transformation $V_ix+b_i$ followed by activation $\phi$. Therefore, the Jacobian matrix can be computed by 
$J_i(x) = \phi'\circ V_i$. The $j$-th row of $E[|J_i|]$ is
\begin{equation}
	\mathbb{E}[|J_i|]_{j,*} = \int |\phi'(x)|t_{i, j}(x)dx \circ |V_i|_{j,*}
\end{equation}
where the subscript $j,*$ means the $j$-th row of the matrix. 

The above process describes the propagation of approximation error through the $i$-th hidden layer. 
Applying the propagation calculation recursively from the first hidden layer to the output layer, we have the following result.

\begin{theorem}[Approximation error propagation]
	\label{theo:approxerror}
	Given a deep neural network $f:\mathbb{R}^d \rightarrow \mathbb{R}^c$ and a linear approximation neural network $g: \mathbb{R}^d \rightarrow \mathbb{R}^c$ learned from $f$. The approximation error 
	\begin{equation}
	\label{eq:approxerror:egf}
	\mathcal{E}(g; f) = \frac{1}{c}\sum (|V_o|\ \mathbb{E}[|r_L|]),
	\end{equation}
	where, for $i = 2,\ldots, L$, 
	\begin{equation}
	\label{eq:er}
	\mathbb{E}[|r_i|] \leq \mathbb{E}[e_i] + \mathbb{E}[|J_i|]\mathbb{E}[|r_{i-1}|] + \mathbb{E}[|\hat{\epsilon_i}|]
	\end{equation} 
	and $\mathbb{E}[|r_1|]  = \mathbb{E}[e_1]$.
\end{theorem}

Based on Theorem~\ref{theo:approxerror}, \nop{ignoring the Taylor approximation error $\epsilon_i$ and} expanding Eq.~(\ref{eq:er}), we have
\begin{equation}
	\label{eq:approxerror:erl}
	\mathbb{E}(|r_L|) \approx \sum_{i=1}^L \prod_{q = L}^{i+1} \mathbb{E}[|J_q|] (\mathbb{E}[e_i] + \mathbb{E}[|\hat{\epsilon_i}|])
\end{equation}
Plugging Eq.~(\ref{eq:approxerror:erl}) into Eq.~(\ref{eq:approxerror:egf}), the model approximation error $\mathcal{E}(g;f)$ can be rewritten in terms of $E[e_{i,j}]$, that is,
\begin{equation}
	\vspace{-1mm}
	\small
	\label{eq:e:linearfunc}
	\mathcal{E}(g; f) = \sum_{i,j} \underbrace{\frac{1}{c}\sum(\ |V_o|\prod_{q=L}^{i+1} \mathbb{E}[|J_q|]\ )_{*, j}}_{w^{(e)}_{i,j}} (\mathbb{E}[e_{i,j}] + \mathbb{E}[|\hat{\epsilon}_{i,j}|])
	\vspace{-1mm}
\end{equation}
here $\sum(\cdot)_{*,j}$ sums up the $j$-th columns, $w^{(e)}_{i,j}$ is the amplification coefficient of $\mathbb{E}[e_{i,j}]$ reflecting its amplification in the subsequent layers to influence the output, and is independent from the approximation of $g$ and is only determined by $f$. 
When $\mathcal{E}(g;f)$ is small and the approximation of $g$ is very close to $f$, the error $\hat{\epsilon_i}$ can be ignored, $\mathcal{E}(g; f)$ is roughly considered a linear combination of $\mathbb{E}[e_{i,j}]$ with amplification coefficient $w^{(e)}_{i,j}$.

\subsection{Approximation Algorithm}
\label{sec:approxalgo}

We use the LANN with the smallest number of linear regions that meets the requirement of approximation degree, which measured by approximation error $\mathcal{E}(g;f)$, to assess the complexity of a model.
Unfortunately, the actual number of linear regions corresponding to data manifold~\cite{bishop2006pattern} in the input-space is unknown. To tackle the challenge, we notice that a piecewise linear activation function with $k$ subfunctions contributes $k-1$ hyperplanes to the input-space partition~\cite{montufar2014number}.
Motivated by this, we propose to minimize the number of hyperplanes under the expectation of minimizing the number of linear regions.  
Formally, under a requirement of approximation degree $\lambda$, our algorithm learns a LANN model with minimum $K(g)=\sum_{i,j} k_{i,j}$.
\nop{
\begin{equation}
	\min(K(g))\ \ \ \ s.t.\ \ \mathcal{E}(g; f) < \lambda
\end{equation}
}
Before presenting our algorithm, we first introduce how we obtain the PDF $t_{i,j}$ of neuron $\{i,j\}$.

\subsubsection{\textbf{Distribution of activation function}}
In Section~\ref{sec:approxerror}, in order to compute $\mathbb{E}[e_{i,j}]$ and $\mathbb{E}[|J_i|]$, we introduce the probability density function $t_{i,j}$ of neuron $\{i,j\}$. To compute $t_{i,j}$, the distribution of activation function is involved. The distribution of an activation function is how outputs (or inputs) of a neuronal activation function distribute with respect to the data manifold. It is influenced by the parameters of previous layers and the distribution of input data. Since the common curve activation functions are bounded to a small output range, to simplify the calculation, we study the posterior distribution of an activation function~\cite{frey1999variational, ioffe2015batch} instead of the input distribution. 
To estimate the posterior distribution, we use kernel density estimation (KDE)~\cite{silverman2018density} with Gaussian kernel, and use the output of activation function $\phi_{i,j}$ on training dateset as the distributed samples $\{x_1, x_2, \ldots, x_n\}$. we have
$
	t_{i,j} = \frac{1}{nh} \sum_{q=1}^n K(\frac{x - x_q}{h})
$
where the bandwidth $h$ is chosen by the rule-of-thumb estimator~\cite{silverman2018density}. To compute $\mathbb{E}[e_{i,j}]$ and $\mathbb{E}[|J_i|]$, we uniformly sample $n_t$ points $\{\Delta x_1, \ldots, \Delta x_{n_t}\}$ within the output range of $\phi$, where $\Delta x_i - \Delta x_{i-1}= \frac{\phi(\infty) - \phi(-\infty)}{n_t}$. 
We then use the expectation \nop{of the discrete} on these samples as an estimation of $\mathbb{E}[e_{i,j}]$. 
\begin{equation}
	\label{eq:approxerror:eeijdiscrete}
	\mathbb{E}[e_{i,j}] \approx \sum_{q=1}^{n_t} e_{i,j}(\Delta x_q)t_{i,j}(\Delta x_q)
\end{equation} 
The output of $\phi$ is smooth and in small range. Setting large sample size $n_t$ does not lead to obvious improvement in the expectation estimation. In our experiments, we set $n_t=200$.
Notice that $x_q$ is the output of $\phi$. The corresponding input is $\phi^{-1}(x_q)$. Thus, $e_{i,j}(x_q) = |\ell_{i,j}(\phi^{-1}(\Delta x_q)) - \Delta x_q|$. $\mathbb{E}[|J_i|]$ is computed in the same way. 

\subsubsection{\textbf{Piecewise linear approximation of activation}}

\begin{algorithm}[t]
	\caption{nextTangentPoint}
	\label{alg:unitwise:nextcut}
	\small{
		\KwIn{ $\phi$, $\ell$, $t$}
		\KwOut{$p^*$, $\mathbb{E}[e]_-$}
		
		\Begin{    
			$\{\Delta x_1, \ldots, \Delta x_{n_t}\} \leftarrow$ uniformly sampled points\;         
			Compute $\mathbb{E}[e]$ by Eq.~(\ref{eq:approxerror:eeijdiscrete})\;
			
			\For{$\Delta x$ {\bfseries in} $\{\Delta x_1, \ldots, \Delta x_{n_t}\}$}{
				$\ell'_{\Delta x} =$ add tangent line of $\Delta x$ to $\ell$\;
				Compute $\mathbb{E}[e(\ell'_{\Delta x})]$\;
			}
			$\Delta x^* = \arg\min_{\Delta x} \mathbb{E}[e(\ell'_{\Delta x})] $\;
			$\mathbb{E}[e]_{-} = \mathbb{E}[e] - \mathbb{E}[e(\ell'_{\Delta x*})]$\;
	}}
\end{algorithm}

To minimize $K(g)$, the piecewise linear approximation function $\ell_{i,j}$ of an arbitrary neuron $\{i,j\}$ is initialized with a linear function ($k=1$). Then every new subfunction is added to $\ell_{i,j}$ to minimize the value of $\mathbb{E}[e_{i,j}]$. Every subfunction is a tangent line of $\phi$. The initialization is the tangent line at $(0, \phi(0))$, which corresponds to the linear regime of the activation function~\cite{ioffe2015batch}. A new subfunction is added to the \textit{next tangent point} $(p^*, \phi(p^*))$, which is found from the set of uniformly sampled points $\{\Delta x_1, \Delta x_2, \ldots, \Delta x_{n_t}\}$. That is,
\begin{equation}
	p^*_{i,j} = arg\min_p \mathbb{E}[e_{i,j}]_{+p};\ \ \ p\in \{\Delta x_1,  \ldots, \Delta x_{n_t}\}
\end{equation}
where subscript $+p$ means that $\ell_{i,j}$ with additional tangent line of $(p, \phi(p))$ is used in computing $\mathbb{E}[e_{i,j}]$. Algorithm~\ref{alg:unitwise:nextcut} shows the pseudocode of determining the next tangent point. 

\subsubsection{\textbf{Building LANNs}}

To minimize $K(g)$, the algorithm starts with initializing every approximation function $\ell_{i,j}$ with a linear function ($k=1$). Then, we iteratively add a subfunction to the approximation function of a certain neuron to decrease $\mathcal{E}(g;f)$ to the most degree in each step. 

In Eq.~(\ref{eq:e:linearfunc}), when building a LANN, the error $\hat{\epsilon_i}$ cannot be ignored because $\mathbb{E}[e_{i,j}]$ is large. The amplification coefficient $w^{(e)}_{i,j}$ of lower layer is exponentially larger than that of the upper layer. Otherwise, error $\mathbb{E}[\hat{\epsilon}_{i,j}]$ grows exponentially from lower to upper layer. Deriving this formula to get the exact weight of $\mathbb{E}[e_{i,j}]$ is complicated.     
A simple way is to roughly consider each $\mathbb{E}[e_{i,j}]$ to be equally important in the algorithm. Specifically, for a neuron from the first layer, small $\mathbb{E}[e_{i,j}]$ is desired due to a large magnitude of $w^{(e)}_{i,j}$ even through $\mathbb{E}[\hat{\epsilon}_{i,j}]=0$. Another neuron from the last hidden layer, its amplification coefficient $w^{(e)}_{i,j}$ is with the lowest magnitude over all layers but $\mathbb{E}[\hat{\epsilon}_{i,j}]$ is not ignorable and may influence the distribution of neuron status, thus approximation with small $\mathbb{E}[e_{i,j}]$ is desired to decrease the value of $\mathbb{E}[e_{i,j}]$ and $\mathbb{E}[\hat{\epsilon}_{i,j}]$. 

\begin{algorithm}[t]
	\caption{BuildingLANN}
	\label{alg:unitwise}
	
	\small{
		\KwIn{a DNN $f(x)$ with activation function $\phi$; training dataset $D_{tr}$; a set of  activation function distributions $T=\{t_{i,j}\}$; batchsize $b$; approximation degree $\lambda$}
		\KwOut{a LANN model $g$}
		\Begin{
			Initialize $\ell_{i,j}$ in $g$ with linear functions\;
			\For{$i \leftarrow 1$ \KwTo $L$}{
				\For{$j \leftarrow 1$ \KwTo $m_i$}{
					Compute $\mathbb{E}[e_{i,j}]$ by Eq.(\ref{eq:approxerror:eeijdiscrete})\;
					$p^*_{i,j},\ \mathbb{E}[e_{i,j}]_-\leftarrow$nextTangentPoint$(\phi, \ell_{i,j}, t_{i,j})$\;
				}
			}
			\Repeat{$\mathcal{E}(g;f) \leq \lambda$}{
				$N_u = $ select $b$ neurons with maximum $E[e_{i,j}]_-$\;
				\For{every neuron $u \in N_u$}{
					$\ell_u\leftarrow$ add tangent line of $p^*_u$ to $\ell_u$\;
					$\mathbb{E}[e_u] = \mathbb{E}[e_u] - \mathbb{E}[e_u]_-$\;
					$p^*_u, \mathbb{E}[e_u]_-\leftarrow$nextTangentPoint$(\phi,\ell_{i,j}, t_{i,j})$\;
				}
				$\mathcal{E}(g;f) \leftarrow$ approximation error on $D_{tr}$\;
			}
		}
	}
\end{algorithm}

Algorithm~\ref{alg:unitwise} outlines the LANN building algorithm. 
To reduce the calculation times, we set up the batch size $b$ to batch processing a group of neurons.
\nop{
\begin{itemize}
	\item[a)] Initialize $g$ to set every $\ell_{i,j}$ to be a linear function ($k_{i,j}=1$). Find next tangent point $p^*_{i,j}$ for every $\ell_{i,j}$.
	
	\item[b)] Find neuron $\{i^*,j^*\}$
	\begin{equation}
	\{i^*,j^*\} = \arg\max (\mathbb{E}[e_{i,j}] - \mathbb{E}[e_{i,j}]_{+p^*_{i,j}}),
	\end{equation}
	Tangent line of $(p^*_{i^*, j^*}, \phi(p^*_{i^*, j^*}))$ is considered currently the most effective addition. 
	\nop{Adding tangent line of $(p^*_{i^*, j^*}, \phi(p^*_{i^*, j^*}))$ to $\ell_{i^*, j^*}$ reduces $\mathcal{E}(g;f)$ most (Eq.~(\ref{eq:e:linearfunc})).}
	
	\item[c)] Add the new subfunction to $\ell_{i^*, j^*}$. Then update its next tangent point $p^*_{i^*, j^*}$. 
	
	\item[d)] Repeat Steps b) and c) until $\mathcal{E}(g;f) < \lambda$.
\end{itemize}
}
The complexity of the algorithm (Algorithm~\ref{alg:unitwise}) is $O(K(g)n)$. The time cost of the first loop is $O((\sum_{i=1}^{L} m_i) * n_t^2)$. The second loop repeats $(K(g)-\sum_{i=1}^L m_i)$ times, within each loop the computation cost is $O((\sum_{i=1}^{L} m_i)+n_t^2+n)$, where $n_t$ is the sample size  of $\phi$ \nop{number of segments of $t(x)$}, $n$ is the number of instances of $D_{tr}$.

%% file: sections/complexity.tex
\nop{
The last section describes how to build a LANN to the target model so that the approximation satisfies a given approximation degree requirement and has as few linear regions as possible.}
The number of linear regions in LANN reflects how nonlinear, or how complex the function of the target model is. In this section, we propose an upper bound to the number of linear regions\nop{ of a LANN}, then propose the model complexity measure based on the upper bound.

The idea of measuring model complexity using the number of linear regions is common in piecewise linear neural networks~\cite{pascanu2013number,raghu2017expressive, montufar2014number, novak2018sensitivity}.  
We generalize their results to the LANN model, of which the major difference is that, in LANN, each piecewise linear activation function has different form and different number of subfunctions.

\nop{Remind our LANN building algorithm (Algorithm~\ref{alg:unitwise}), under a required approximation degree $\lambda$, the algorithm aims to reach $\lambda$ with as small iteration steps as possible by decreasing $\mathcal{E}(g;f)$ as much as possible in each step. Each iteration step increases 1 to the number of subfunctions of a certain $\ell_{i,j}$. A $\ell$ with $k$ subfunctions is considered contributing $k-1$ hyperplanes to the linear region splitting of input space. In this case, our algorithm aims to approximate the target model with as small linear regions as possible in the way of optimizing the number of hyperplane for linear region splitting.}

\begin{theorem}[Upper bound]
	\label{theo:nregion}
	Given a linear approximation neural network $g:\mathbb{R}^d\rightarrow \mathbb{R}^c$ with $L$ hidden layers. Let $m_i$ be the width of the $i$-th layer and $k_{i,j}$ the number of subfunctions of $\ell_{i,j}$. The number of linear regions of $g$ is  
	upper bounded by 
	\begin{math}
	\prod_{i=1}^{L} (\sum_{j=1}^{m_i} k_{i, j} - m_i + 1)^{d}
	\end{math}.
\end{theorem} 

Please see Appendix~\ref{sec:apdx:proof} for the proof of Theorem~\ref{theo:nregion}. 
This theorem indicates that the number of linear regions is polynomial with respect to layer width and exponential with respect to layer depth. This is consistent with the previous studies on the power of neural networks~\cite{bianchini2014complexity, bengio2011expressive, poole2016exponential, eldan2016power}. Meanwhile, the value of $k$ reflects the nonlinearity of the corresponding neuron according to the status distribution of activation functions. The distribution is influenced by both model parameters and data manifold. Thus, this upper bound reflects the impact of model parameters on complexity. Based on this upper bound, we define the complexity measure.

\begin{definition}[Complexity measure]
	\label{def:complexity}
	\small
	Given a deep neural network $f$ and a linear approximation neural network $g$ learned from $f$ with approximation degree $\lambda$, the $\lambda$-approximation complexity measure of $f$ is
	\begin{equation}
	C(f)_\lambda = d \sum_{i=1}^L \log(\sum_{j=1}^{m_i}k_{i,j} - m_i + 1)
	\end{equation}
\end{definition}
This complexity measure is essentially a simplification of our proposed upper bound by logarithm.
We recommend to select $\lambda$ from the range of $\mathcal{E}$ when $\frac{(\mathcal{E}')^2}{\mathcal{E}''}$ converges to a constant. (Appendix~\ref{sec:apdx:lambda})

\nop{\mc{The complexity measure has two important desirable properties. First, deterministic of complexity measure. For a given deep neural network $f$. Under fixed approximation degree $\lambda$ and fixed distribution of activation functions, the learned linear approximation neural network $g$ is deterministic. The complexity measured by $g$ is deterministic. This is obvious since the LANN building algorithm has no indeterministic steps. }\todo{Second, stability of complexity measure. Under a mild assumption that when approximation error $\mathcal{E}(g;f)$ decreases to a low value the second-order derivative of $\mathcal{E}(g;f)$ is constant. Then for two models $f_1$ and $f_2$, if $C(f_1)_{\lambda 1} < C(f_2)_{\lambda 1}$, for $\lambda 2 < \lambda 1$, $C(f_1)_{\lambda 2} < C(f_2)_{\lambda 2}$.}}

\nop{\begin{claim}[Deterministic of complexity measure]
		For a given deep neural network $f$. Under fixed approximation degree $\lambda$ and fixed distribution of activation functions, the learned linear approximation neural network $g$ is deterministic. The complexity measured by $g$ is deterministic. 
	\end{claim}
	\todo{A proof should be given.}
}

%% file: sections/application.tex
\label{sec:application}

In this section, we take several empirical studies to shed more insights on the complexity measure. First, we investigate various contributions of hidden neurons to model stability. Then, we examine the changing trend of model complexity in the training process. After that, we study the occurrence of overfitting and $L^1$ and $L^2$ regularizations. Finally, we propose two new simple and effective approaches to prevent overfitting. 

\begin{table}[t]
    \caption{Model structure of DNNs in our experiments. }
    \label{tab:network}
    \resizebox{\linewidth}{!}{
        \begin{tabular}{rccc}
            \toprule
            & Sec~\ref{sec:application:neuron} & Sec~\ref{sec:application:training} & Sec~\ref{sec:application:reg}, \ref{sec:application:newreg} \\        
            \midrule
            MOON & - & -& L3M(32,128,16)$_T$ \\
            MNIST & L3M300$_S$ & L3M100$_T$, L6M100$_T$, L3M200$_T$ & - \\
            CIFAR & L3M300$_T$ & L3M200$_T$, L6M200$_T$, L3M400$_T$ & L3M(768,256,128)$_T$ \\
            \bottomrule
        \end{tabular}
    }
\end{table}

\begin{figure}[t]
    \begin{center}
        \subfigure[MNIST]{
            \includegraphics[width = 0.43\linewidth]{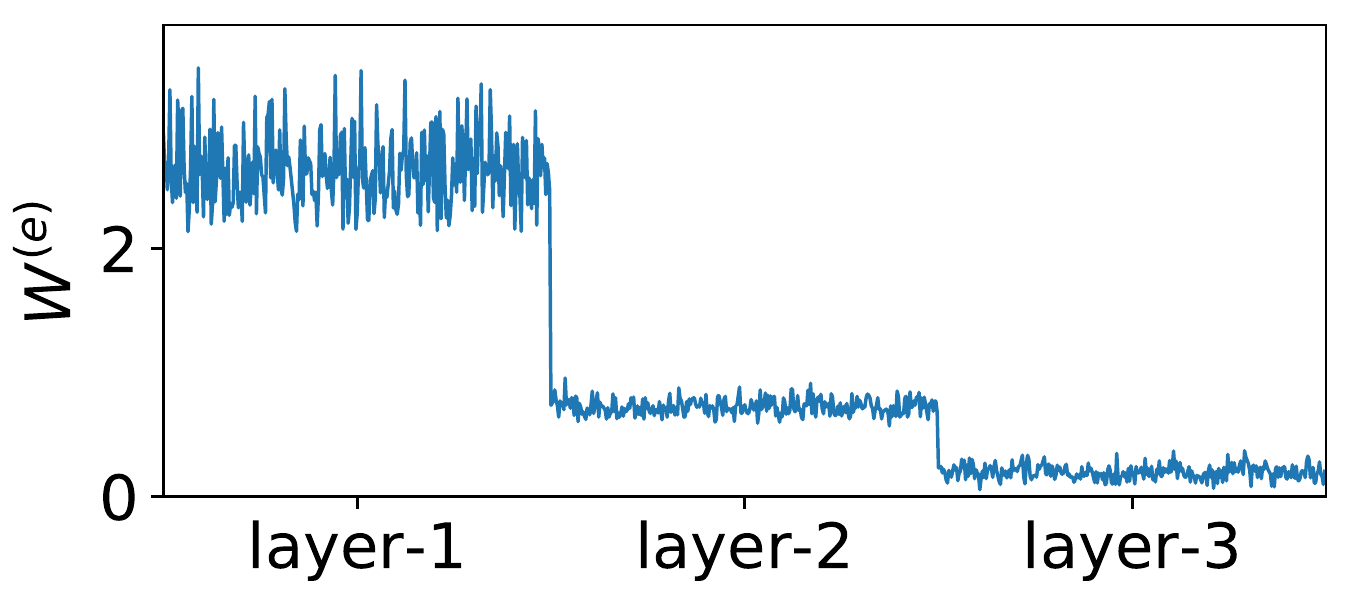}
        }
        \hspace{.15in}
        \subfigure[CIFAR]{
            \includegraphics[width=0.43\linewidth]{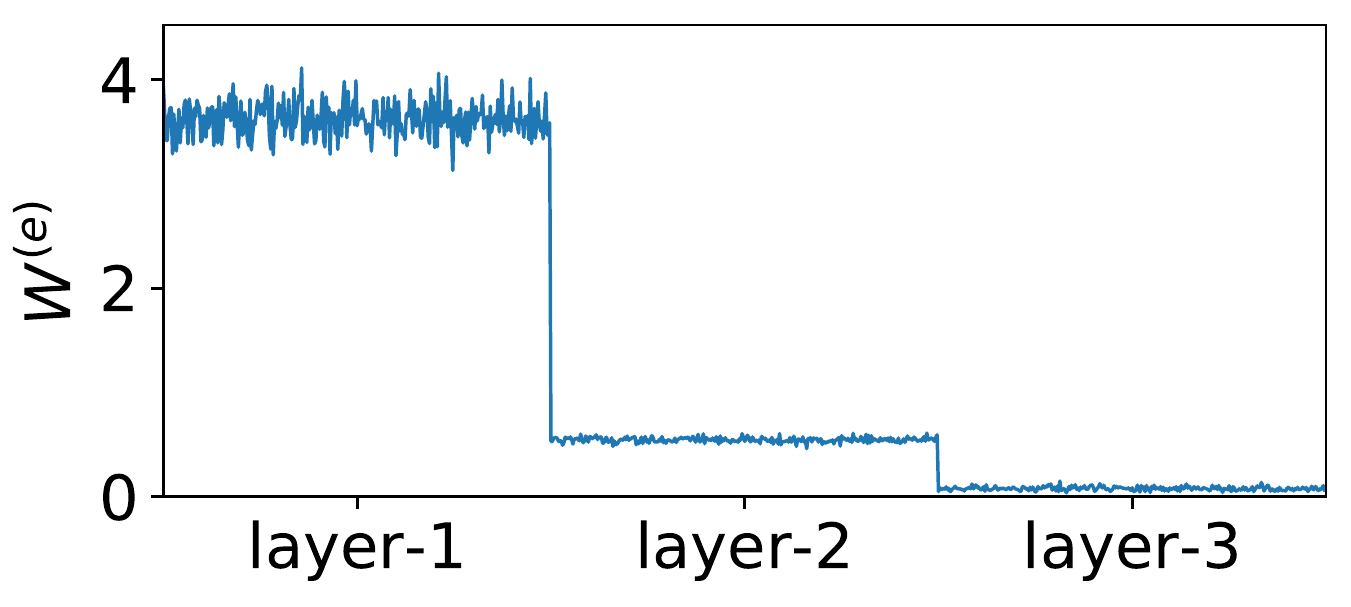}    
        }
        \caption{Amplification coefficient of every neuron. }
        \label{fig:neuronimportance:we}
    \end{center}
\end{figure}
\begin{figure}[t]
    \begin{center}
        \subfigure[MNIST]{
            \includegraphics[width=0.45\linewidth]{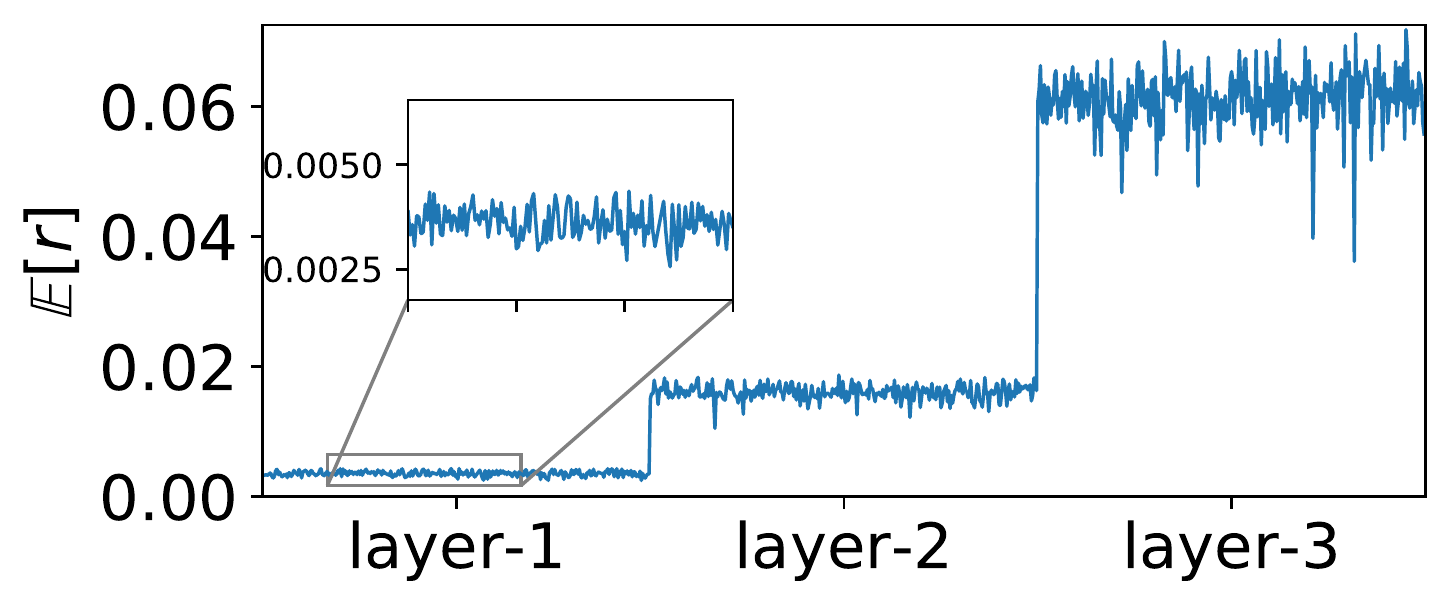}
        }
        \hspace{.1in}
        \subfigure[CIFAR]{
            \includegraphics[width=0.45\linewidth]{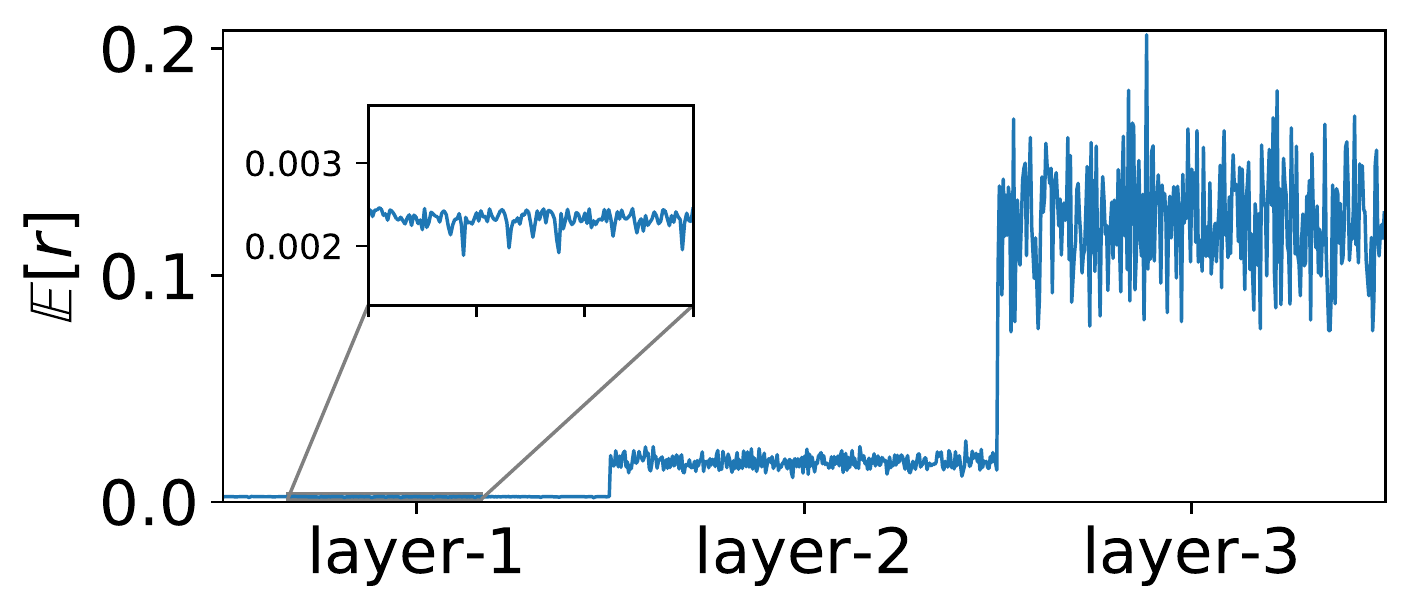}            
        }
        \caption{Layerwise error accumulation ($\lambda =0.1$). }
        \label{fig:neuronimportance:er}
    \end{center}
\end{figure}

Our experiments and evaluations are conducted on both synthetic (Two-Moons\footnote{The synthetic dataset is generated by sklearn.datasets.make\_moons API. }) and real-world datasets (MNIST~\cite{lecun1998gradient}, CIFAR-10~\cite{krizhevsky2009learning}). To demonstrate that the reliability of the complexity measure does not depend on model structures, we design multiple model structures. We use $\lambda=0.1$ for complexity measure in all experiments, which sits in our suggested range for all models we used. 
Table~\ref{tab:network} summarizes the model structures we used, where L3 indicates the network is with 3 hidden layers, M300 means each layer contains 300 neurons while M(32,128,16) means that the first, second, and third layers contain 32, 128, and 16 neurons, respectively. Subscripts $S$ and $T$ stand for the activation functions Sigmoid and Tanh, respectively.

\subsection{Hidden Neurons and Stability}
\label{sec:application:neuron}

As discussed in Section~\ref{sec:approxerror}, the amplification coefficient $w^{(e)}_{i,j}$~(Eq.~\ref{eq:e:linearfunc}) is defined by the multiplication of $\mathbb{E}[J_p]$ through subsequent layers. $w^{(e)}_{i,j}$ measures the magnification effect of the perturbation on neuron $\{i,j\}$ in subsequent layers. In other words, the amplification coefficient reflect the effect of a neuron on model stability.
Figure~\ref{fig:neuronimportance:we} visualizes amplification coefficients of trained models on the MNIST and CIFAR datasets, showing that neurons from the lower layers have greater amplification factors. To exclude the influence of variant layer widths, each layer of the models has the equal width.  

Besides amplification coefficient, we also visualize $\mathbb{E}[r_i]$, the error accumulation of all previous layers. According to our analysis, $\mathbb{E}[r_i]$ is expected to have the opposite trend with $w^{(e)}_{i,j}$: $\mathbb{E}[r]$ of upper layers is expected to be exponentially larger than lower layers. Figure~\ref{fig:neuronimportance:er} shows error accumulation $\mathbb{E}[r_i]$ on the same models.

\begin{figure}
    \begin{center}
        \subfigure[MNIST]{
            \includegraphics[width=0.43\linewidth]{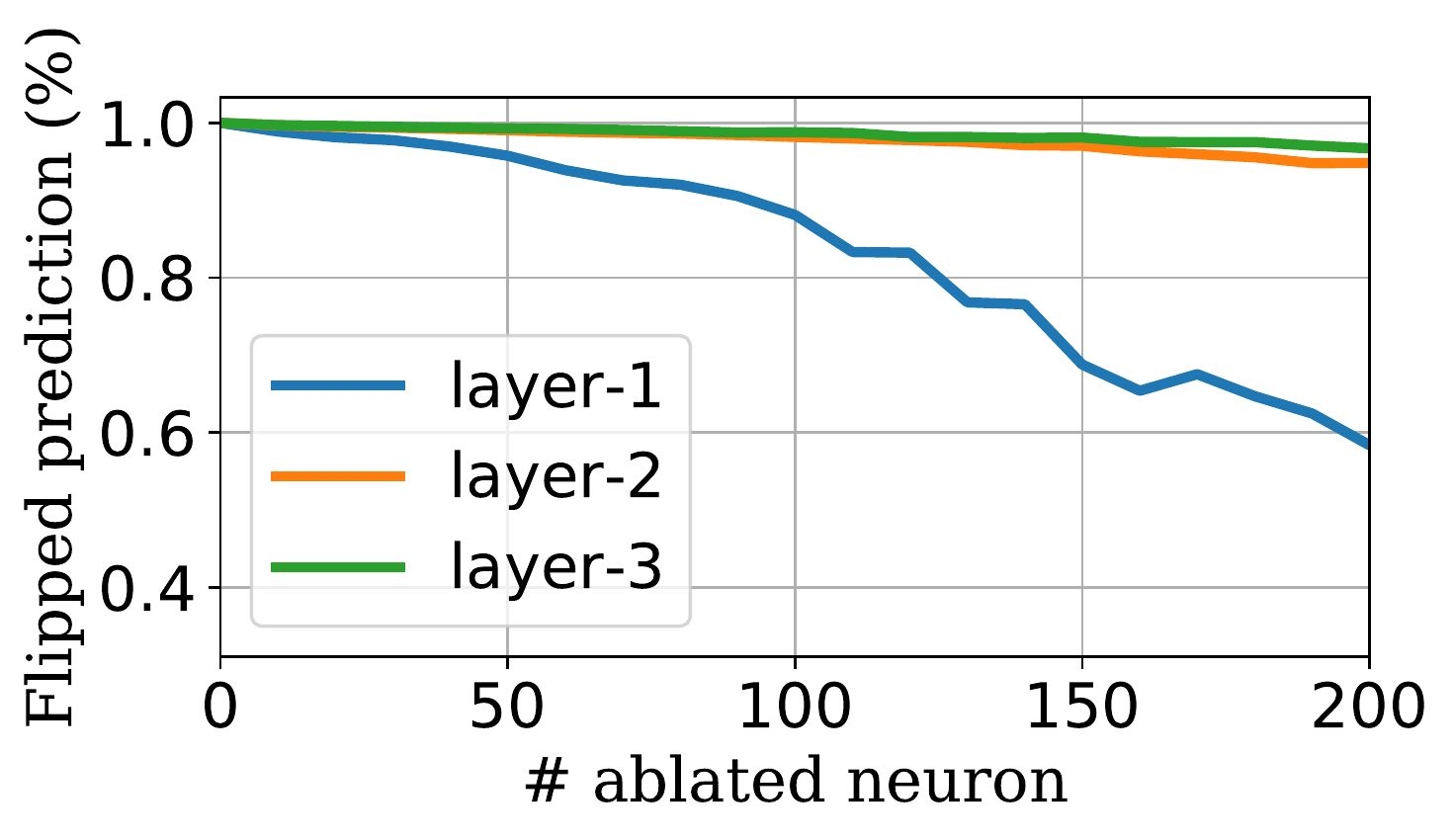}
        }
        \hspace{.2in}
        \subfigure[CIFAR]{
            \includegraphics[width=0.43\linewidth]{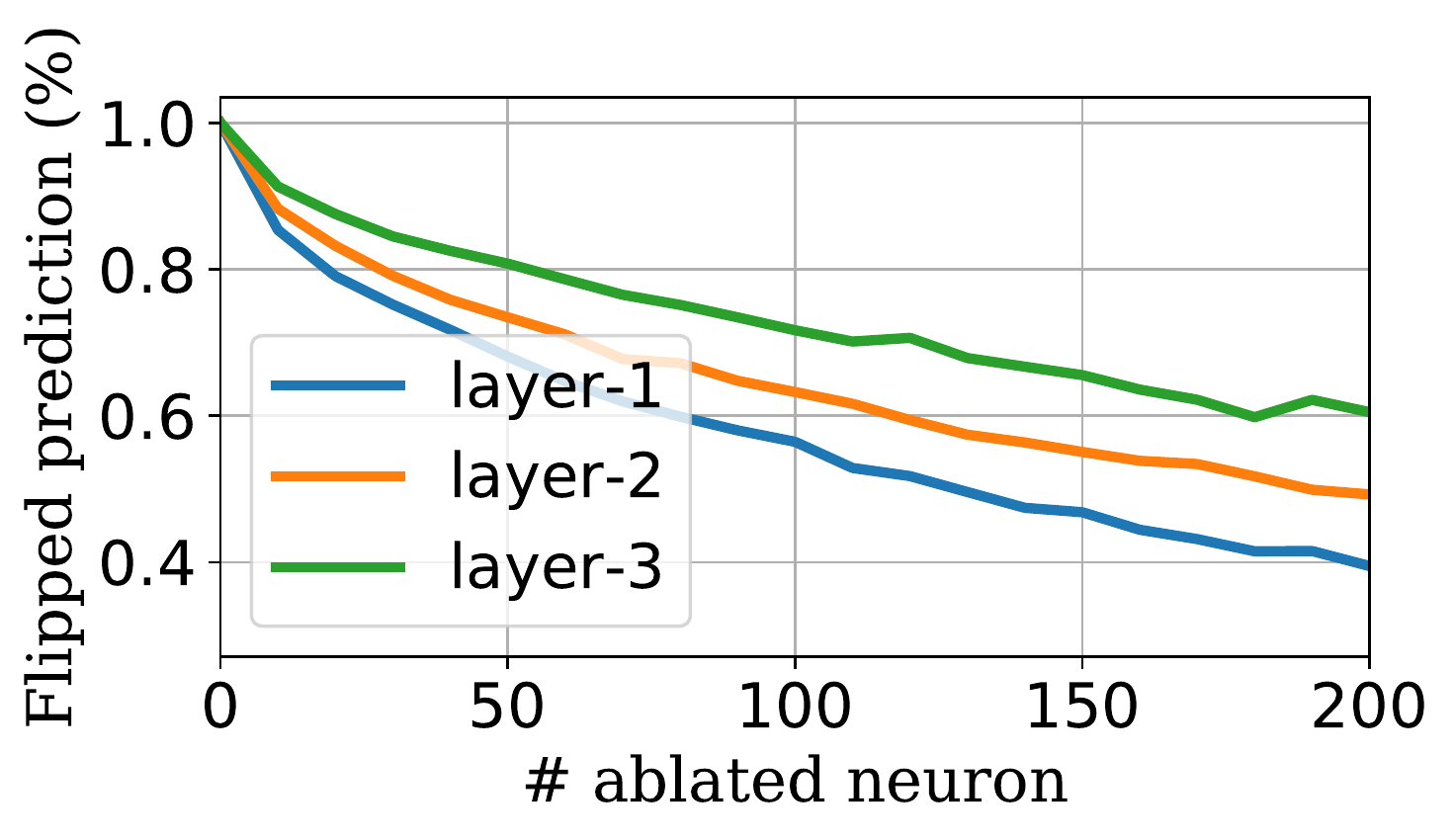}            
        }
        \caption{Percentage of flipped prediction labels after random neuron ablation. }    
        \label{fig:neuronimportance:ablate}    
    \end{center}
\end{figure}

To verify that a small perturbation at a lower layer can cause greater influence on the model outputs than at a upper layer, we randomly ablate neurons (i.e., fixing the neuron output to 0) from one layer of a well-trained model and observe the number of instances whose prediction labels are consequently flipped. The results of ablating different layers are shown in Figure~\ref{fig:neuronimportance:ablate}. 

\subsection{Complexity in Training}
\label{sec:application:training}

\begin{figure}[t]
    \begin{center}
        \subfigure[MNIST]{
            \includegraphics[width=0.47\linewidth]{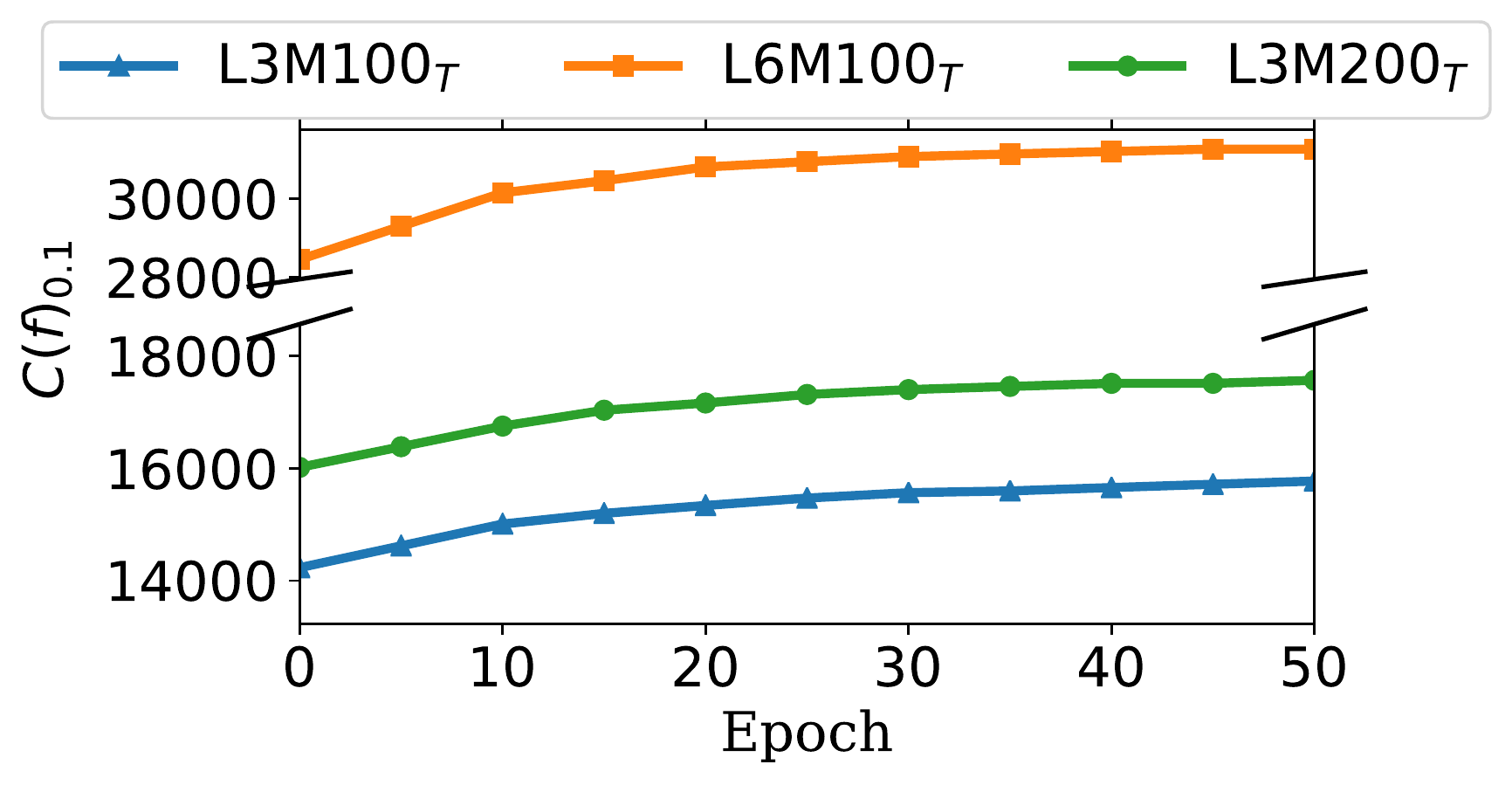}
        }
        \subfigure[CIFAR]{
            \includegraphics[width=0.47\linewidth]{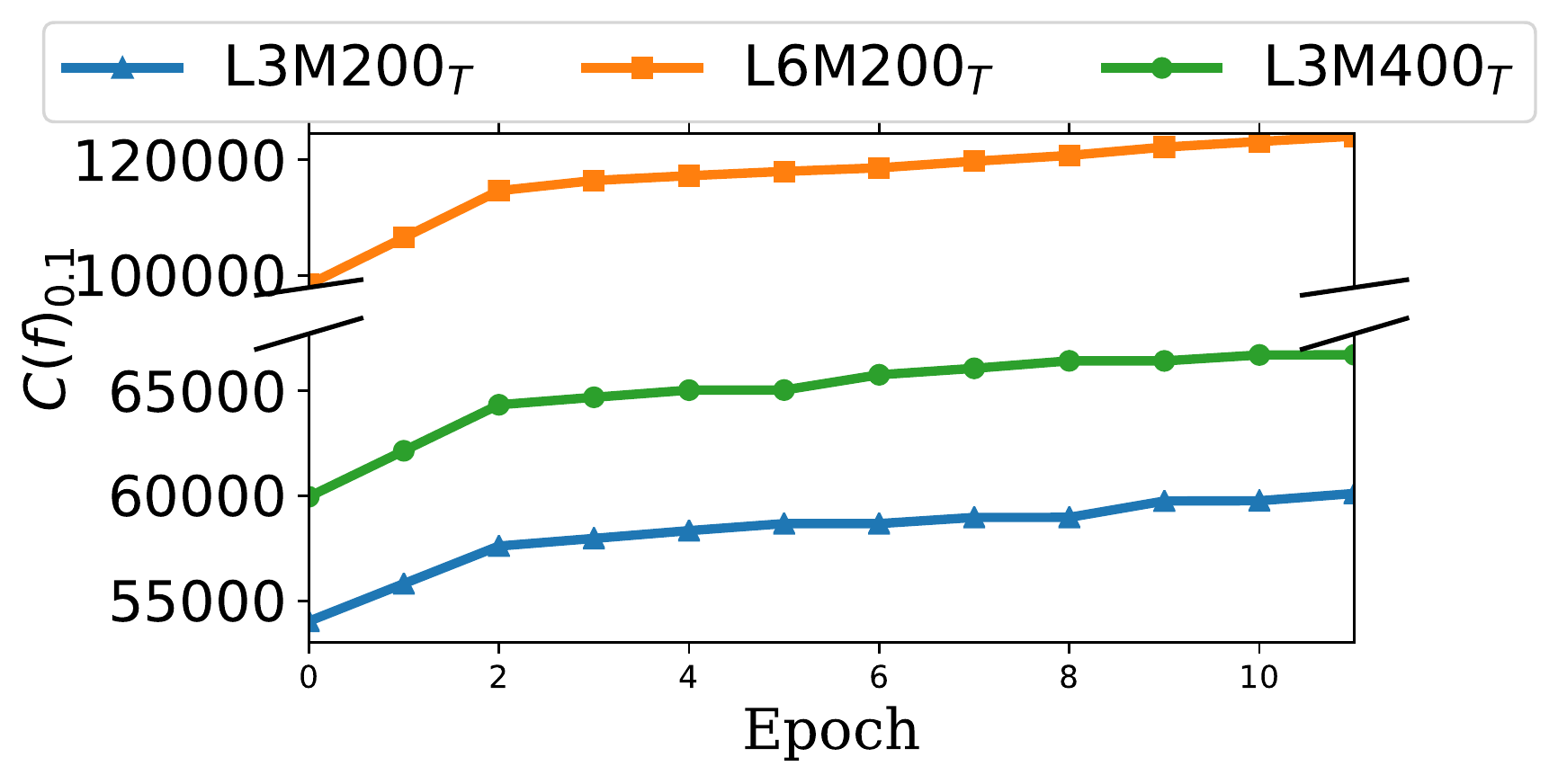}
        }
        \caption{Changing trend of complexity measure in training process of three models on MNIST dataset. }
        \label{fig:complexintraining}
    \end{center}
\end{figure}

In this experiment, we investigate the trend of changes in model complexity in the training process. 
Figure~\ref{fig:complexintraining} shows the periodically-recorded model complexity measure during training based on the 0.1-approximation complexity measure $C(f)_{0.1}$. From this figure, we can observe the soaring model complexity along with the training, which indicates that the learned deep neural networks become increasingly complicated. Figure~\ref{fig:complexintraining}  sheds light on how the model structure influences the complexity measure. Particularly, it is clear to see that increases in both width and depth can increase the model complexity. Furthermore, with the same number of neurons, the complexity of a deep and narrow model (L6M100$_T$ on MNIST, L6M200$_T$ on CIFAR) is much higher than a shallow and wide one (L3M200$_T$ on MNIST, L3M400$_T$ on CIFAR). This agrees with the existing studies on the effectiveness of width and depth of DNNs \cite{montufar2014number, pascanu2013number, bengio2011expressive, eldan2016power}.

\subsection{Overfitting and Complexity}
\label{sec:application:reg}

\begin{figure}[t]
    \begin{center}
        \includegraphics[width=0.6\linewidth]{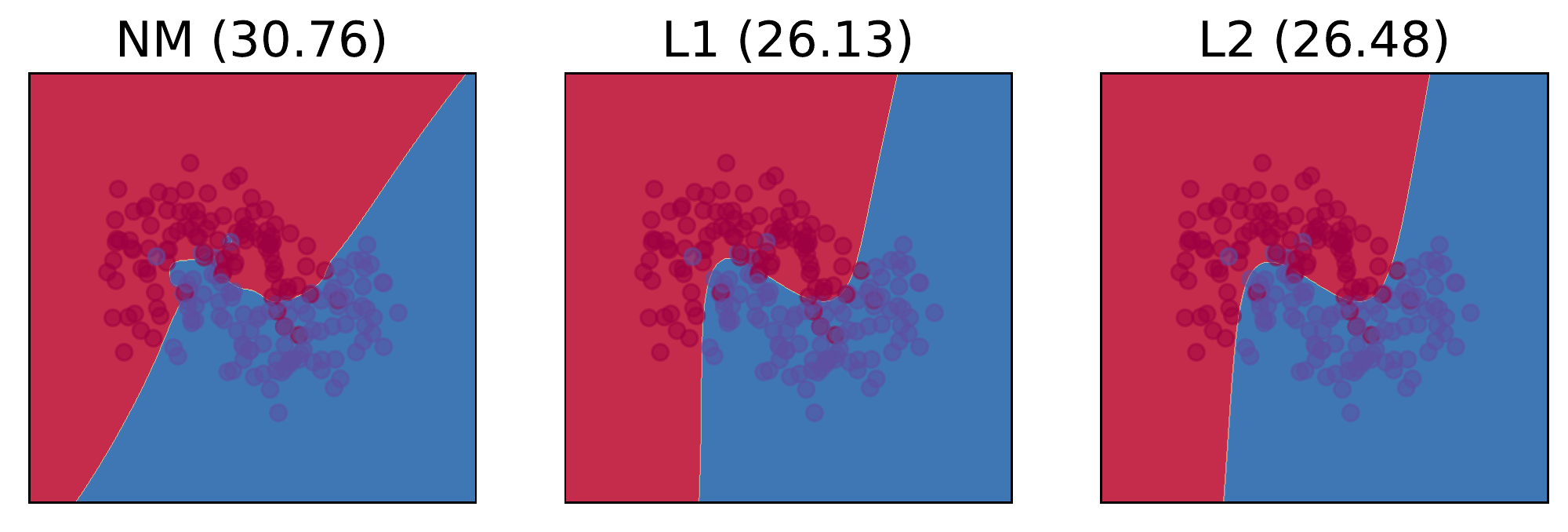}
        \caption{Decision boundaries of models trained on MOON dataset. NM, L1, L2 are short for normal train, train with $L^1$, $L^2$ regularization respectively. In brakets are the value of complexity measure $C(f)_{0.1}$. }
        \label{fig:regularization:moon}
    \end{center}
\end{figure}
\begin{figure}[t]
    \begin{center}
        \subfigure[Overfitting]{
            \includegraphics[width=0.45\linewidth]{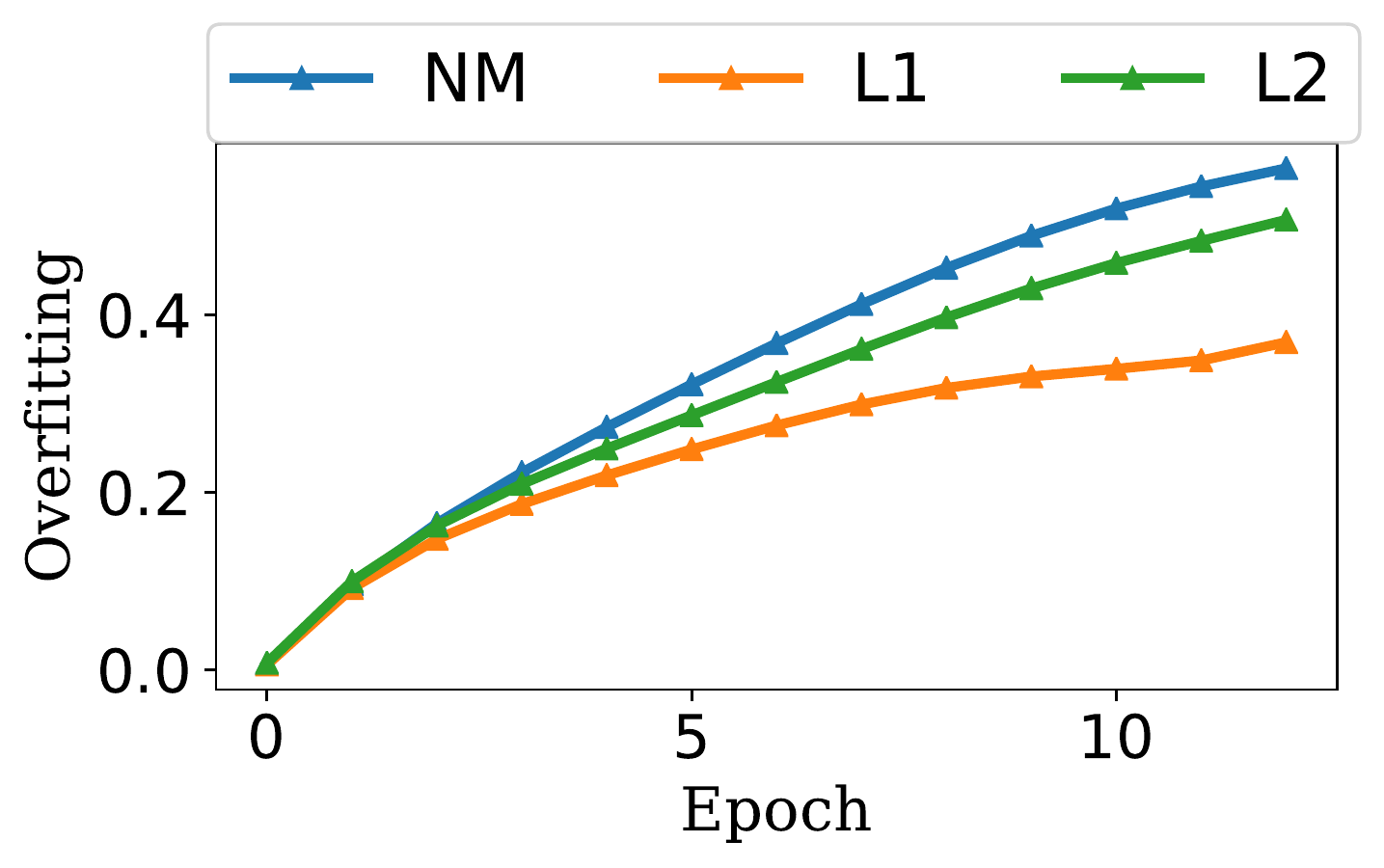}
            \label{fig:regularization:cifar:of}
        }
        \subfigure[Complexity measure]{
            \includegraphics[width=0.49\linewidth]{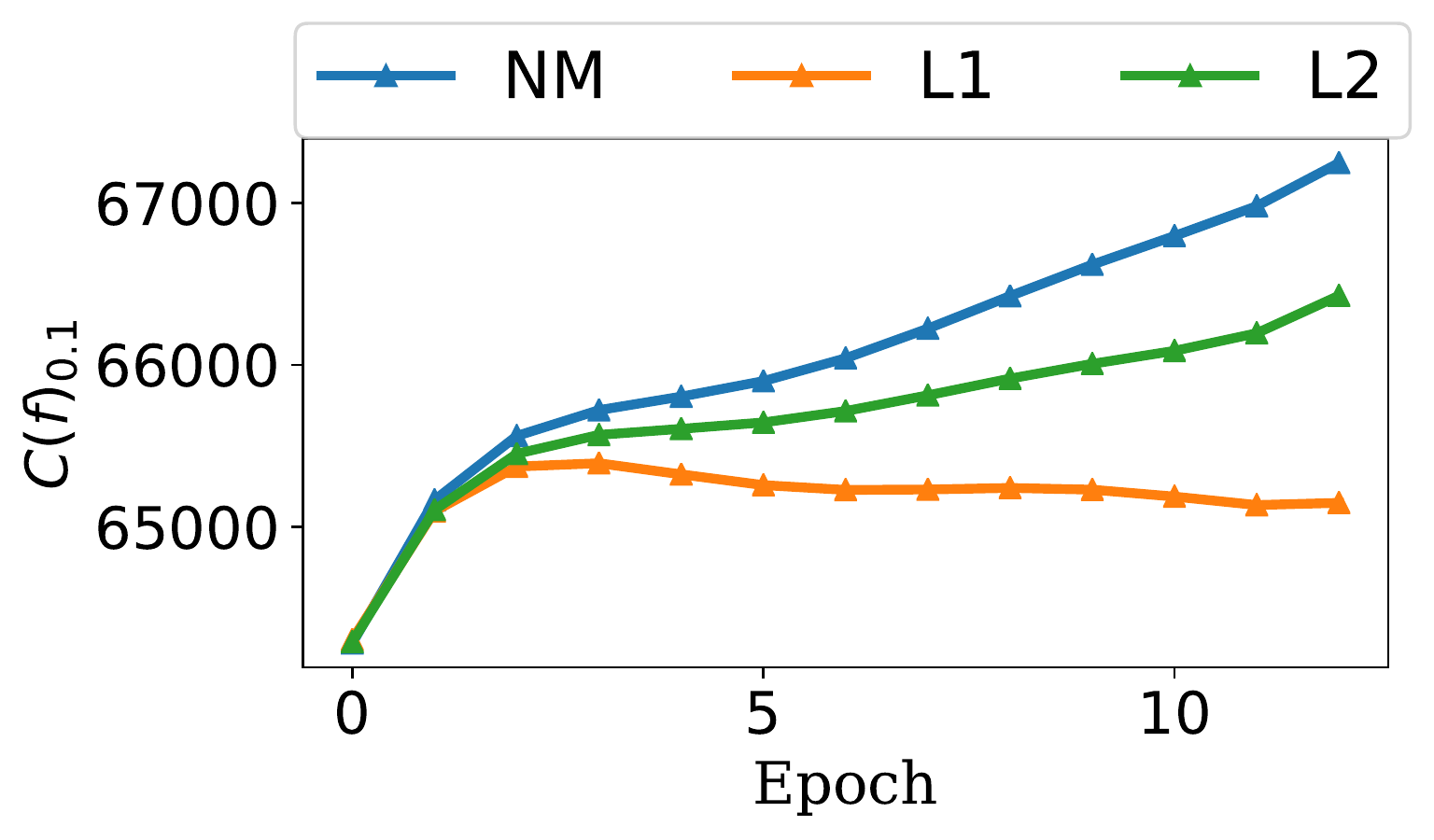}
            \label{fig:regularization:cifar:c}
        }
        \caption{Complexity measure during training of CIFAR dataset. Weight penalties are $1e-4$ and $1e-3$ for $L^1$ and $L^2$ regularizations, respectively. }
        \label{fig:regularization:cifar}
    \end{center}
\end{figure}

The complexity measure through LANNs can be used to understand overfitting. Overfitting usually occurs when training a model that is unnecessarily flexible~\cite{hawkins2004problem}. Due to the high flexibility and strong ability to accommodate curvilinear relationships, deep neural networks suffer from overfitting if they are learned by maximizing the performance on the training set rather than discovering the patterns which can be generalized to new data~\cite{goodfellow2016deep}. Previous studies~\cite{hawkins2004problem} show that an overfitting model is more complex than not overfitting ones. This idea is intuitively demonstrated by the polynomial fit example in Figure~\ref{fig:intro:overfitting}.

Regularization is an effective approach to prevent overfitting, by adding regularizer to the loss function, especially $L^1$ and $L^2$ regularization~\cite{goodfellow2016deep}. $L^1$ regularization results in a more sparse model, and $L^2$ regularization results in a model with small weight parameters. A natural hypothesis is that these regularization approaches can succeed in restricting the model complexity. To verify this, we train deep models on the MOON dataset with and without regularization. After 2,000 training epochs, their decision boundaries and complexity measure $C(f)_{0.1}$ are shown in Figure~\ref{fig:regularization:moon}. The results demonstrate the effectiveness of $L^1$ and $L^2$ regularizations in preventing overfitting and constraining increase of the model complexity. 

We also measure model complexity during the training process, after each epoch of CIFAR, with or without $L^1$ and $L^2$ regularizations. The results are shown in Figure~\ref{fig:regularization:cifar}. Figure~\ref{fig:regularization:cifar:of} is the overfitting degree measured by $(Accuracy_{train} - Accuracy_{test})$, Figure~\ref{fig:regularization:cifar:c} is the corresponding complexity measure $C(f)_{0.1}$. The results verify the conjecture that $L^1\ and\ L^2$ regularizations constrain the increase of model complexity.

\subsection{New Approaches for Preventing Overfitting}
\label{sec:application:newreg}

Motivated by the well-observed significant correlation between the occurrence of overfitting and the increasing model complexity, we propose two approaches to prevent overfitting by directly suppressing the rising trend of the model complexity during training. 

\subsubsection{Neuron Pruning} From the definition of complexity~(Def.~\ref{def:complexity}), we know that constraining model complexity $C(f)_{\lambda}$, i.e., restraining the variable $k_{i,j}$ for each neuron, is equivalent to constraining the non-linearity of the distribution of a neuron. Thus, we can periodically prune neurons with a maximum value of $\mathbb{E}[|t|]$, after each training epoch.  This is inspired by the fact that a larger value of $\mathbb{E}[t]$ implies the higher probability that the distribution $t$ is located at the nonlinear range and therefore requires a larger $k$. Pruning neurons with a potentially large degree of non-linearity can effectively suppress the rising of model complexity. At the same time, pruning a limited number of neurons unlikely significantly decreases the model performance. Practical results demonstrate that this approach, though simple, is quite efficient and effective. 

\subsubsection{Customized $L^1$ Regularization} This is to give customized coefficient to every column of weight matrix $V_i (i=1,\ldots, L)$ when doing $L^1$ regularization. Each column corresponds to a specific neuron and with coefficient:
\begin{equation}
    a_{i,j} = \mathbb{E}[|\phi_{i,j}'|] = \int |\phi'(x)|t_{i,j}(x)dx
\end{equation} 
One explanation is that $a_{i,j}$ equals to the expectation of first-order derivative of $\phi_{i,j}$. With a larger value of $\mathbb{E}[|\phi'_{i,j}|]$, the distribution $t_{i,j}$ is with a higher probability located at the linear range of the activation function ($0=argmax_x \phi'(x)$). The customized $L^1$ approach assigns larger sparse penalty weights to more linearly distributed neurons. The neurons with more nonlinear distributions can maintain their expressive power. Another view to understand this approach is to using Eq.~(\ref{eq:approxerror:erl}), $a_{i,*}|V_i| = \mathbb{E}[|J_i|]$. That is, the formulation of customized $L^1$ can be interpreted as the constraint of $E[|J|]$, which will obviously result in smaller $\mathcal{E}(g;f)$ as well as smaller $C(f)$. Customized $L^1$ is more flexible than the normal $L^1$ regularization, thus behaves better with large penalty weight.

\begin{table}[t]
    \caption{Complexity measure and number of linear regions of MOON. }
    \label{tab:mooncomplexity}
    \begin{tabular}{rccccc}
        \toprule
        & NM & PR & C-L1 & L1 & L2 \\
        \midrule
        $C(f)_{0.1}$ & 31.17 & 25.02 & 25.11 & 25.78 & 26.55 \\
        \# Regions & 45,772 & 182 & 356 & 382 & 545 \\
        \bottomrule
    \end{tabular}
\end{table}
\begin{figure}[t]
    \begin{center}
        \includegraphics[width=\linewidth]{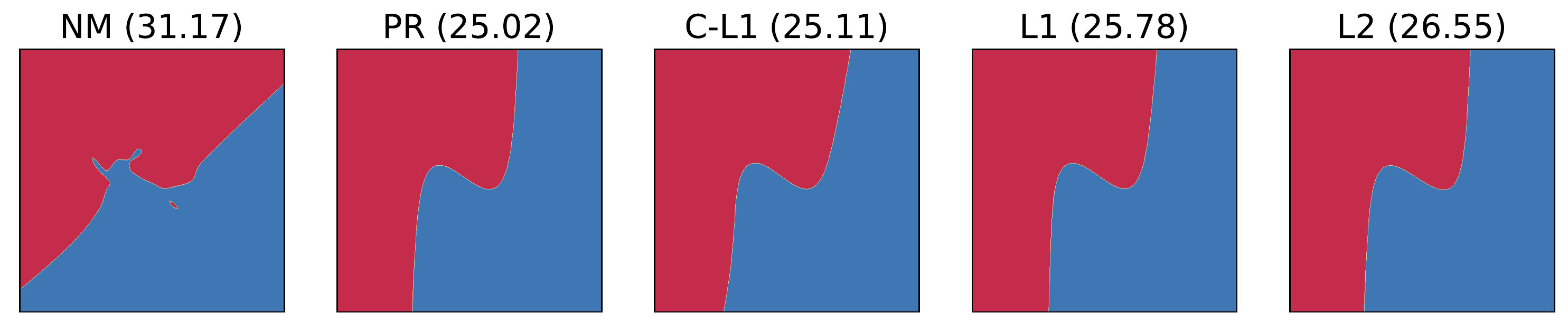}
        \caption{Decision boundaries of models trained with different regularization methods on MOON dataset. PR, C-L1 are short for training with neuron prunning, with customized $L^1$ regularization. }
        \label{fig:regularization:newmoon}
    \end{center}
\end{figure}

\begin{figure}[t]
	\begin{center}
		\subfigure[Overfitting]{
			\includegraphics[width=0.45\linewidth]{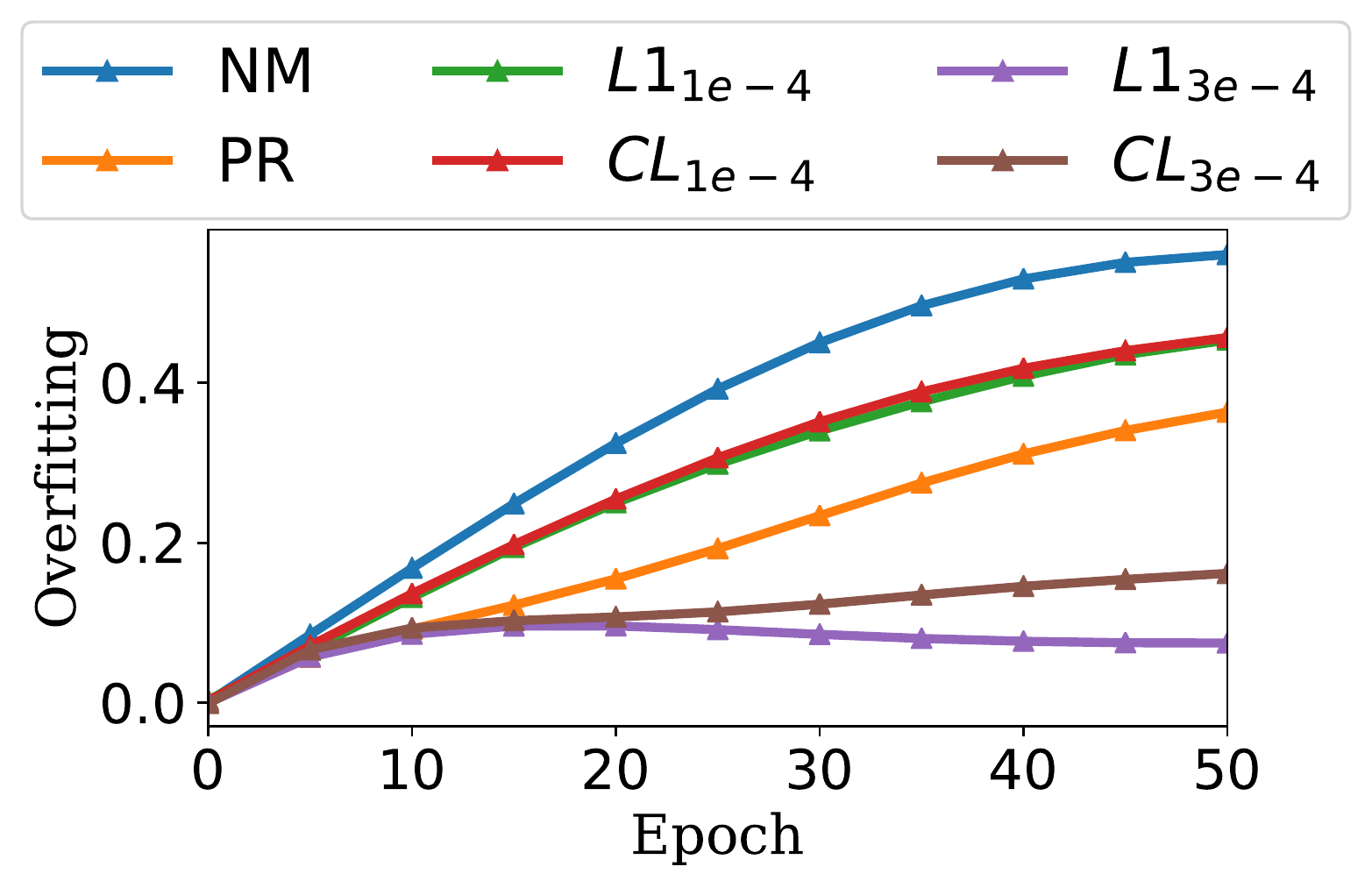}
			\label{fig:regularization:newcifar:of}
		}
		\subfigure[Complexity measure]{
			\includegraphics[width=0.47\linewidth]{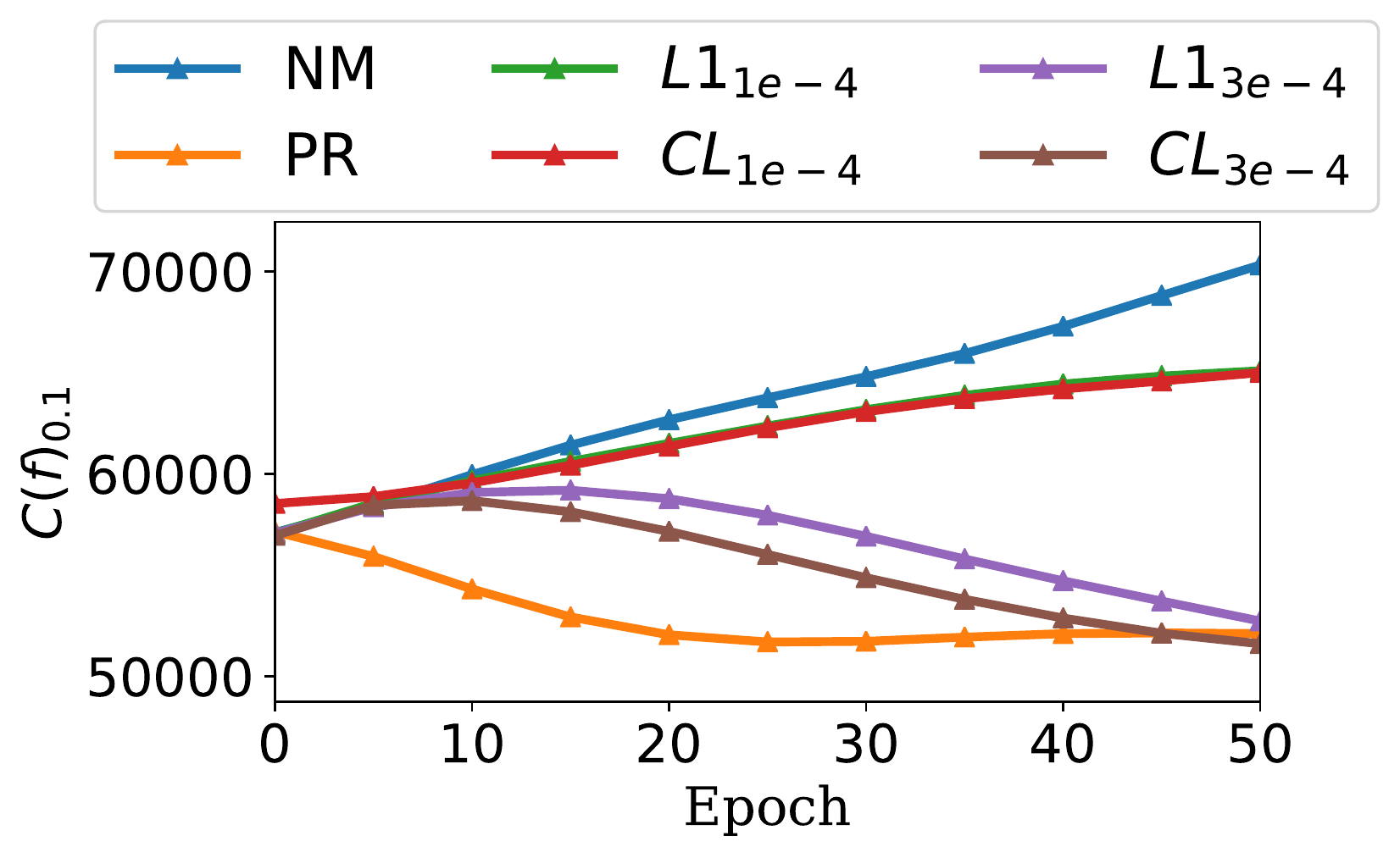}
			\label{fig:regularization:newcifar:c}
		}
		\caption{Degree of overfitting and complexity measure in training process of CIFAR dataset. }
		\label{fig:complexity:newcifar}
	\end{center}
\end{figure}

Figure~\ref{fig:regularization:newmoon} compares the respective decision boundaries of the models trained with different regularization approaches on the MOON dataset. Table~\ref{tab:mooncomplexity} records the corresponding complexity measure and the number of split linear regions over the input space. Figure~\ref{fig:complexity:newcifar} shows the overfitting and complexity measures in the training process of models on CIFAR. 
In our experiments, the neuron pruning percentage set to $5\%$. 
These figures demonstrate that neuron pruning can constrain overfitting and model complexity, and still retain satisfactory model performance. 
We scale the customized $L^1$ coefficient $a_{i,j}$ to $\frac{a_{L1}}{\mathbb{E}[a_{i,j}]}a_{i,j}$
so that its mean value is equal to the penalty weight of $L^1$, denoted by $a_{L1}$.
Our results shows that, with a small penalty weight, the customized $L^1$ approach behaves close to normal $L^1$. With a large penalty weight, the performance of $L^1$ model is affected, test accuracy decrease by $3\%$. The customized $L^1$ approach retains the performance (Appendix~\ref{sec:apdx:experiment}). 

%% file: sections/conclusion.tex
In this paper, we develope a complexity measure for deep \nop{fully-connected} neural networks with curve activation functions. 
{
Particularly, we first propose the linear approximation neural network (LANN), a piecewise linear framework, to both approximate a given DNN model to a required approximation degree and minimize the number of resulting linear regions. 
After providing an upper bound to the number of linear regions formed by LANNs, we define the complexity measure facilitated by the upper bound. 
To examine the effectiveness of the complexity measure, we conduct empirical analysis, which demonstrated the positive correlation between the occurrence of overfitting and the growth of model complexity during training. 
In the view of our complexity measure, further analysis revealed that $L^1, L^2$ regularizations indeed suppress the increase of model complexity. Based on this discovery, we finally proposed two approaches to prevent overfitting through directly constraining model complexity: neuron pruning and customized $L^1$ regularization.
}
There are several future directions, including generalizing the usage of our proposed linear approximation neural network to other network architectures (i.e. CNN, RNN).

\nop{Our proposed complexity measure applies to any deep fully connected neural networks with curve activation functions. For example, a Batch Norm~\cite{ioffe2015batch} layer $\gamma \frac{x - \mu}{\sqrt{\sigma^2 + \epsilon}} + \beta$ influences by contribute a coefficient $\gamma / \sigma$ to the error propogation equation, our proposed approach is still applicable. }

%% file: sections/appendix.tex
\section{Proof and Discussions}

\subsection{Proof of Theorem~\ref{theo:nregion}}
\label{sec:apdx:proof}

\begin{proof}
	First of all, according to~\cite{montufar2014number, raghu2017expressive, pascanu2013number}, the total number of linear regions divided by $k$ hyperplanes in the input space $\mathbb{R}^d$ is upper bounded by $\sum_{i=0}^{d}\tbinom{k}{i}$, whose upper bound can be obtained using binomial theorem: 
	\begin{equation}
		\label{eq:bino}
		\sum_{i=0}^{d}\tbinom{k}{i} \leq (k+1)^d
	\end{equation}
	. 	
		
	Now consider the first hidden layer $h'_1$ of a LANN model. \nop{What is LApprox?}
	A piecewise linear function consisting of $k_{i,j}$ subfunctions contributes $k_{i,j}-1$ hyperplanes to the input space splitting. The first layer $h'_1$ contains $m_1$ neurons, with $j$-th neuron consisting of $k_{1,j}$ subfunctions. So $h'_1$ contributes $\sum_{j=1}^{m_1} (k_{1,j}-1)$ hyperplanes to the input space $\mathbb{R}^d$ splitting, and divides $\mathbb{R}^d$ into linear regions with upper bound (Eq.~\ref{eq:bino}): 
	\begin{equation}
		(\sum_{j=1}^{m_1} k_{1,j} -m_1 + 1)^d
	\end{equation}
	
	Now move to the second hidden layer $h'_2$. For each linear region divided by the first layer, it can be divided by the hyperplanes of $h'_2$ to at most ($\sum_{j=1}^{m_2}k_{2,j} - m_2 + 1)^d$ smaller regions. 
	
	Thus, the total number of linear regions generated by $h'_1, h'_2$ is at most 
	\begin{equation}
		(\sum_{j=1}^{m_1} k_{1,j} -m_1 + 1)^d * (\sum_{j=1}^{m_2}k_{2,j} - m_2 + 1)^d
	\end{equation} 
	. 
	
	Recursively do this calculation until the last hidden layer $h'_L$. Finally, the number of linear regions divided by $g$ is at most
	\begin{equation}
		\prod_{i=1}^{L} (\sum_{j=1}^{m_i} k_{i, j} - m_{i} + 1)^{d}
	\end{equation}
\end{proof}

\subsection{Suggested Range of $\lambda$}
\label{sec:apdx:lambda}

In this section we provide a suggestion of the range of $\lambda$ when using LANN for complexity measure. A suitable value of $\lambda$ makes the complexity measure trustworthy and stable. When the value of $\lambda$ is large, the measure may be unstable and unable to reflect the real complexity. It seems small value of $\lambda$ is prefered, however small value calls for higher cost to construct the LANN approximation. And how small should $\lambda$ be? Based on analyzing the curve of approximation error, we provide an empeircal range. 

\begin{figure}[t]
	\subfigure[Approximation error $\mathcal{E}$]{
		\includegraphics[width=0.4\linewidth]{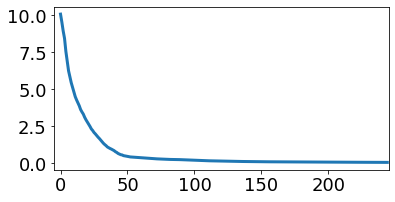}
		\label{fig:appendix:ae}
	}
	\subfigure[Approximation gain $k$]{
		\includegraphics[width=0.4\linewidth]{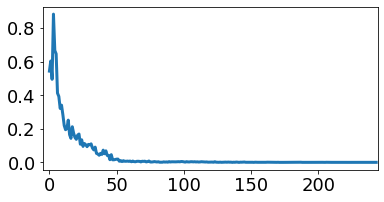}
		\label{fig:appendix:ag}
	}
	\subfigure[Second-order derivative $a$]{
		\includegraphics[width=0.4\linewidth]{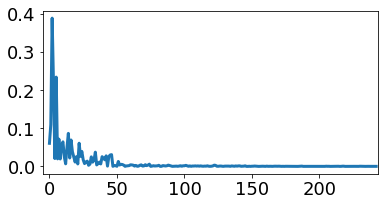}
		\label{fig:appendix:ddae}
	}
	\subfigure[$k^2/a$]{
		\includegraphics[width=0.4\linewidth]{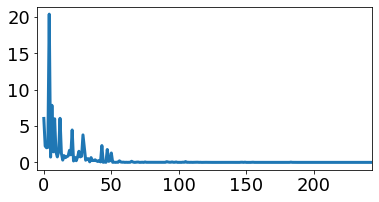}
		\label{fig:appendix:k2a}
	}
	\caption{Changing trend of approximation error $\mathcal{E}$, approximation gain $k$, $a$ which is the second-order derivative of $\mathcal{E}$, and $k^2/a$ computed from $k$ and $a$. }	
	\label{fig:appendix:a}
\end{figure}

We first analyze the curve of approximation error in several aspects. Approximation error $\mathcal{E}$ is the optimization object in building LANN algorithm (Algorithm~\ref{alg:unitwise}), so obviously it goes decreasing during training epochs~(Figure~\ref{fig:appendix:ae}). Meanwhile, the absolute of first-order derivative of $\mathcal{E}$, which represents the contribution of current epoch's operation to the decrease of apporixmation error $\mathcal{E}$, is called \emph{approximation gain} here, and denoted by $k$. Our algorithm ensures that, at any time $k$ is expected to be larger than all remaining possible operations. 
Figure~\ref{fig:appendix:ag} shows the curve of approximation gain. Because we ignore the error $\hat{\epsilon}$ in the algorithm, the curve of approximation gain in practice has a small range of jitter, but the decreasing trend can be guaranteed. 
We also consider the derivative of $k$, formally the absolute of second-order derivative of approximation error $\mathcal{E}$, denoted by $a$. The second-order derivative $a$ reflects the changing trend of the approximation gain $k$. It is easy to prove that, the trend of $a$ goes decrease with training epoch increases: If not, after a finite number of epochs we have $k=0$. But in fact, since $\mathcal{E}$ will never decrease to 0, operation of each epoch brings non-zero influence to $\mathcal{E}$, thus $k$ will not be 0. Figure~\ref{fig:appendix:ddae} shows the change trend of $a$. 

\begin{figure}[t]
	\subfigure[Approximation error $\mathcal{E}$]{
		\includegraphics[width=0.4\linewidth]{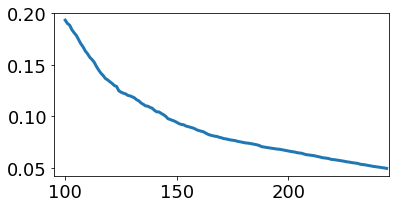}
		\label{fig:appendix:aes}
	}
	\subfigure[Approximation gain $k$]{
		\includegraphics[width=0.4\linewidth]{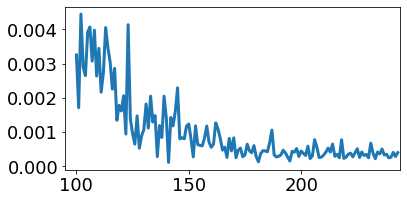}
		\label{fig:appendix:ags}
	}
	\subfigure[Second-order derivative $a$]{
		\includegraphics[width=0.4\linewidth]{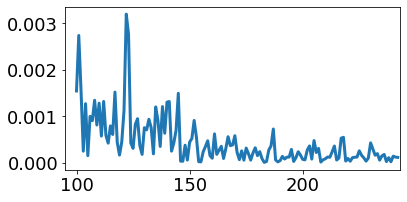}
		\label{fig:appendix:ddaes}
	}
	\subfigure[$k^2/a$]{
		\includegraphics[width=0.4\linewidth]{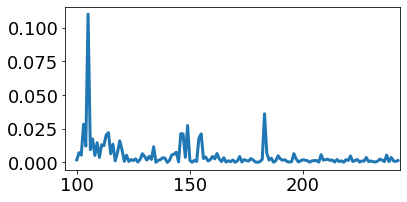}
		\label{fig:appendix:k2as}
	}
	\caption{Changing trend of approximation error $\mathcal{E}$, approximation gain $k$, $a$ which is the second-order derivative of $\mathcal{E}$, and $k^2/a$ computed from $k$ and $a$. Here we enlarge second half, after 100 epoches of Figure~\ref{fig:appendix:a}. }	
	\label{fig:appendix:as}
\end{figure}

See from Figure~\ref{fig:appendix:a}, the changing trends of $\mathcal{E}$,  $k$ and $a$ are close to each other. The trend decreases quickly at the beginning then gradually flatten to convergence. This agrees with our algorithm design. After $\mathcal{E}$ goes flatten, the following relationships are established: $k\rightarrow 0, a\rightarrow 0$, $k \neq 0,  a \neq 0$, $a << k$. 

Suppose there is an epoch $t_0$ in the flatten region of $\mathcal{E}$, $k, a$ are its first-order, second-order derivative. We show changing trends of flatten regions in Figure~\ref{fig:appendix:as}. According to Figure~\ref{fig:appendix:a} and the above analysis, the curve after $t_0$ is basically stable. We estimate the total gain of approximation error that can be brought by remaining epochs. Suppose there exists a $n$ that after $n$ epochs from $t_0$, $k$ goes 0. Then the gain of remaining epochs are the gain of the next $n$ epochs. Suppose $a$ is constant, $n=k/a$. the gain of remaining epochs is estimated by $kn-an^2/2 = k^2/2a$. 

We analyze $k$ and $a$ from the view of the remaining gain estimation. In practice, $k$ and $a$ keep decrease. If $k$ and $a$ goes stable and with very close decreasing trend, the estimation of remaining gain of $t_0$ should be close to the estimation of epochs around $t_0$. 
Suppose the above condition is true, we have:  $k^2/a \approx (k+a)^2/(a+a') \Rightarrow k/a \approx a/a'$, where $a'$ is the derivative of $a$. This is, the downward trend of $k$ and $a$ are basically similar, and $a' << a << k << 1$ is true. 

As a result, $k^2/a$ of an epoch almost equalling to the calculated value of its neighbors demonstrates that, the derivative of $k$ and $a$ are almost the same. The gain of remaining epoches are expected to be relatively stable, each afterward epoch will not bring much influence to the value of $\mathcal{E}$. In this case, the $\mathcal{E}$ is relatively stable.

The conclusion is, for the construction of a LANN based on a specific target model, $\lambda < \lambda_0$ is suggested where $\lambda_0$ is the starting point of $k^2/a$ converging to a constant.  

\begin{figure}[t]
	\subfigure[Approximation error $\mathcal{E}$]{
		\includegraphics[width = 0.4\linewidth]{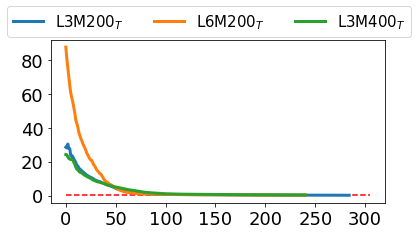}
	}
	\subfigure[$k^2/a$]{
		\includegraphics[width=0.4\linewidth]{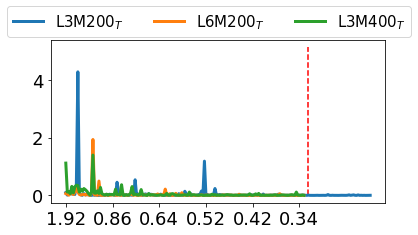}
	}
	\caption{Verify the rationality of $\lambda = 0.1$ for three models trained on CIFAR: L3M200$_T$, L6M200$_T$, L3M400$_T$. Left figure shows the curve of approximation errors of three models. Right figure shows the value $k^2/a$ in the area nearby $0.1$. Here $x$ axis is the corresponding approximation error. }
	\label{fig:appendix:mnist}
\end{figure}

For the comparable of two LANNs, find such $\lambda$ which satisfying $\lambda < min(\lambda_{0,a}, \lambda_{0,b})$ and $k_a(\lambda) \approx k_b(\lambda)$.  This to some degree ensures the stability of complexity measure of the target model, the estimated gain of remaining epochs of two LANNs are almost similar.

In practical experiments, the value of $k^2/a$ is used to check if the value of $\lambda$ is reasonable. In our experiments, we choose a uniform $\lambda = 0.1$ and verify its rationality. From our experimental results, it seems for relatively simple network (e.g. 3 layers, hundreds of width), $\lambda \leq 0.12$ is good enough since the $k^2/a$ goes convergence. In Figure~\ref{fig:appendix:mnist} we show the changing trends on the CIFAR to demontrate that $\lambda=0.1$ is a reasonable value in our experiments.

\subsection{More Experimental Results}
\label{sec:apdx:experiment}

\subsubsection{{Extension of Section~\ref{sec:application:newreg}}}
In Section~\ref{sec:application:newreg}, we report that customized $L^1$ regularization is more flexible than normal $L^1$ regularization, such that behaves better with large weight penalty. We indicate that customized $L^1$ maintains the prediction performance on the CIFAR test dataset while $L^1$ is about $3\%$ lower. Below in Figure~\ref{fig:appendix:reg} we show the corresponding prediction accuracy on training and test dataset. 

\begin{figure}[t]
    \includegraphics[width=0.43\linewidth]{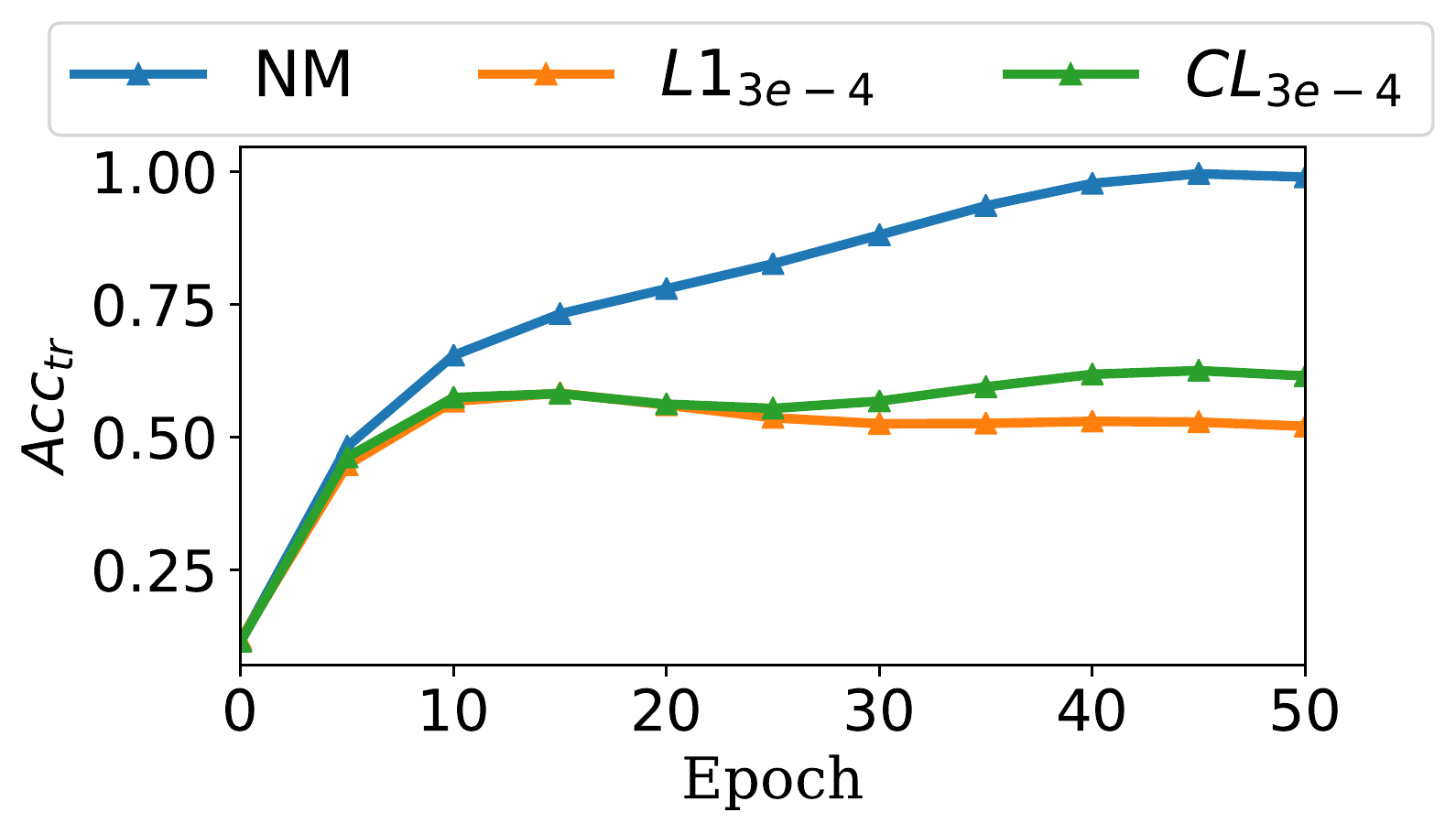}
    ~
	\includegraphics[width=0.43\linewidth]{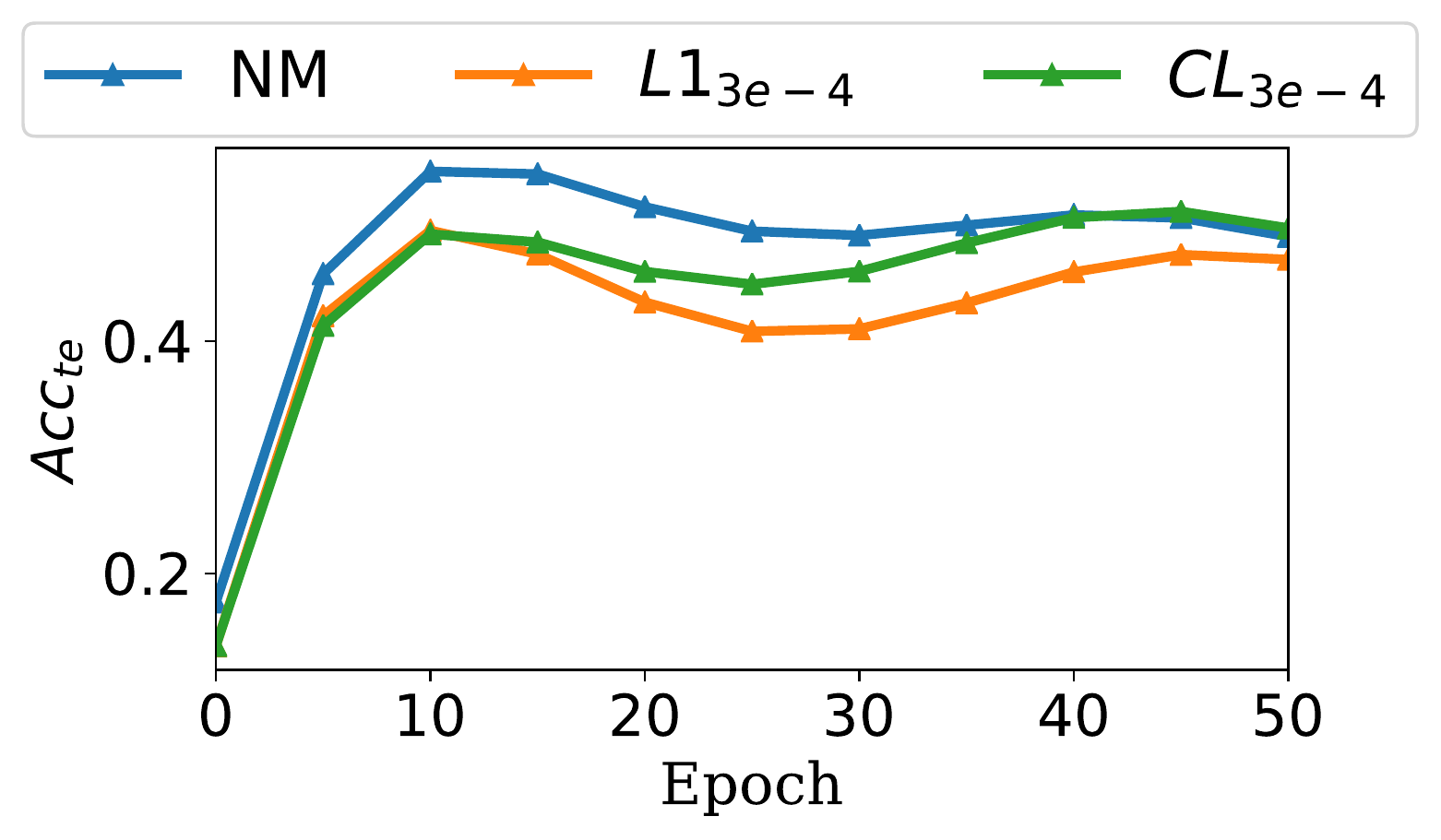}
	\caption{Left shows the accuracy on the CIFAR training dataset, the right one shows the accuracy on the CIFAR test dataset. Both in the training process. }
	\label{fig:appendix:reg}
\end{figure}

\subsubsection{{Complexity Measure is Data Insensitive}}

\begin{table}[t]
	\caption{Compare approximation error on training dataset and test dataset.  }
	\label{tab:apdx:network}
	\begin{tabular}{cccc}
		\toprule
		Dataset & Model & $\mathcal{E}_{train}$ & $\mathcal{E}_{test}$ \\		
		\midrule
		MNIST &  L3M100$_T$ & 0.0999 & 0.0988 \\
		MNIST & L6M100$_T$ & 0.0979 & 0.0971 \\
		MNIST & L3M200$_T$ & 0.0911 & 0.0907 \\
		MNIST & L3M300$_S$ & 0.0944 & 0.0942 \\
		CIFAR & L3M300$_T$ & 0.0989 & 0.0977 \\
		CIFAR & L3M200$_T$ & 0.0979 & 0.0984 \\
		CIFAR & L6M200$_T$ & 0.0973 & 0.0970 \\
		CIFAR & L3M400$_T$ & 0.0984 & 0.0976 \\
		CIFAR & L3M(768,256,128)$_T$ & 0.0970 & 0.0979 \\
		\bottomrule
	\end{tabular}
\end{table}

To verify if our complexity measure by LANN is data sensitive, we measure the approximation error of LANNs on test dataset. Below in Table~\ref{tab:apdx:network} we compare approximation errors on training dataset (the dataset used to build LANNs) and test dataset. The results show that LANNs achieve very close approximation error on training and test dataset, which demonstrates that our complexity measure is data dependence but data insensitive.

%% file: main.bbl

\begin{thebibliography}{35}


\ifx \showCODEN    \undefined \def \showCODEN     #1{\unskip}     \fi
\ifx \showDOI      \undefined \def \showDOI       #1{#1}\fi
\ifx \showISBNx    \undefined \def \showISBNx     #1{\unskip}     \fi
\ifx \showISBNxiii \undefined \def \showISBNxiii  #1{\unskip}     \fi
\ifx \showISSN     \undefined \def \showISSN      #1{\unskip}     \fi
\ifx \showLCCN     \undefined \def \showLCCN      #1{\unskip}     \fi
\ifx \shownote     \undefined \def \shownote      #1{#1}          \fi
\ifx \showarticletitle \undefined \def \showarticletitle #1{#1}   \fi
\ifx \showURL      \undefined \def \showURL       {\relax}        \fi
\providecommand\bibfield[2]{#2}
\providecommand\bibinfo[2]{#2}
\providecommand\natexlab[1]{#1}
\providecommand\showeprint[2][]{arXiv:#2}

\bibitem[\protect\citeauthoryear{Barron}{Barron}{1993}]%
        {barron1993universal}
\bibfield{author}{\bibinfo{person}{Andrew~R Barron}.}
  \bibinfo{year}{1993}\natexlab{}.
\newblock \showarticletitle{Universal approximation bounds for superpositions
  of a sigmoidal function}.
\newblock \bibinfo{journal}{\emph{IEEE Transactions on Information theory}}
  \bibinfo{volume}{39}, \bibinfo{number}{3} (\bibinfo{year}{1993}),
  \bibinfo{pages}{930--945}.
\newblock


\bibitem[\protect\citeauthoryear{Bengio and Delalleau}{Bengio and
  Delalleau}{2011}]%
        {bengio2011expressive}
\bibfield{author}{\bibinfo{person}{Yoshua Bengio} {and}
  \bibinfo{person}{Olivier Delalleau}.} \bibinfo{year}{2011}\natexlab{}.
\newblock \showarticletitle{On the expressive power of deep architectures}. In
  \bibinfo{booktitle}{\emph{International Conference on Algorithmic Learning
  Theory}}. Springer, \bibinfo{pages}{18--36}.
\newblock


\bibitem[\protect\citeauthoryear{Bianchini and Scarselli}{Bianchini and
  Scarselli}{2014}]%
        {bianchini2014complexity}
\bibfield{author}{\bibinfo{person}{Monica Bianchini} {and}
  \bibinfo{person}{Franco Scarselli}.} \bibinfo{year}{2014}\natexlab{}.
\newblock \showarticletitle{On the complexity of shallow and deep neural
  network classifiers.}. In \bibinfo{booktitle}{\emph{ESANN}}.
\newblock


\bibitem[\protect\citeauthoryear{Bishop}{Bishop}{2006}]%
        {bishop2006pattern}
\bibfield{author}{\bibinfo{person}{Christopher~M Bishop}.}
  \bibinfo{year}{2006}\natexlab{}.
\newblock \bibinfo{booktitle}{\emph{Pattern recognition and machine learning}}.
\newblock \bibinfo{publisher}{springer}.
\newblock


\bibitem[\protect\citeauthoryear{Chiu, Sainath, Wu, Prabhavalkar,
  et~al\mbox{.}}{Chiu et~al\mbox{.}}{2018}]%
        {chiu2018state}
\bibfield{author}{\bibinfo{person}{Chung-Cheng Chiu}, \bibinfo{person}{Tara~N
  Sainath}, \bibinfo{person}{Yonghui Wu}, \bibinfo{person}{Rohit Prabhavalkar},
  {et~al\mbox{.}}} \bibinfo{year}{2018}\natexlab{}.
\newblock \showarticletitle{State-of-the-art speech recognition with
  sequence-to-sequence models}. In \bibinfo{booktitle}{\emph{2018 IEEE
  ICASSP}}. IEEE, \bibinfo{pages}{4774--4778}.
\newblock


\bibitem[\protect\citeauthoryear{Cohen, Sharir, and Shashua}{Cohen
  et~al\mbox{.}}{2016}]%
        {cohen2016expressive}
\bibfield{author}{\bibinfo{person}{Nadav Cohen}, \bibinfo{person}{Or Sharir},
  {and} \bibinfo{person}{Amnon Shashua}.} \bibinfo{year}{2016}\natexlab{}.
\newblock \showarticletitle{On the expressive power of deep learning: A tensor
  analysis}. In \bibinfo{booktitle}{\emph{Conference on Learning Theory}}.
  \bibinfo{pages}{698--728}.
\newblock


\bibitem[\protect\citeauthoryear{Cybenko}{Cybenko}{1989}]%
        {cybenko1989approximation}
\bibfield{author}{\bibinfo{person}{George Cybenko}.}
  \bibinfo{year}{1989}\natexlab{}.
\newblock \showarticletitle{Approximation by superpositions of a sigmoidal
  function}.
\newblock \bibinfo{journal}{\emph{Mathematics of control, signals and systems}}
  \bibinfo{volume}{2}, \bibinfo{number}{4} (\bibinfo{year}{1989}),
  \bibinfo{pages}{303--314}.
\newblock


\bibitem[\protect\citeauthoryear{Delalleau and Bengio}{Delalleau and
  Bengio}{2011}]%
        {delalleau2011shallow}
\bibfield{author}{\bibinfo{person}{Olivier Delalleau} {and}
  \bibinfo{person}{Yoshua Bengio}.} \bibinfo{year}{2011}\natexlab{}.
\newblock \showarticletitle{Shallow vs. deep sum-product networks}. In
  \bibinfo{booktitle}{\emph{Advances in NIPS}}. \bibinfo{pages}{666--674}.
\newblock


\bibitem[\protect\citeauthoryear{Ding, Zhang, Liu, and Duan}{Ding
  et~al\mbox{.}}{2015}]%
        {ding2015deep}
\bibfield{author}{\bibinfo{person}{Xiao Ding}, \bibinfo{person}{Yue Zhang},
  \bibinfo{person}{Ting Liu}, {and} \bibinfo{person}{Junwen Duan}.}
  \bibinfo{year}{2015}\natexlab{}.
\newblock \showarticletitle{Deep learning for event-driven stock prediction}.
  In \bibinfo{booktitle}{\emph{Proceeding of the 24th IJCAI}}.
\newblock


\bibitem[\protect\citeauthoryear{Du and Lee}{Du and Lee}{2018}]%
        {du2018power}
\bibfield{author}{\bibinfo{person}{Simon~S Du} {and} \bibinfo{person}{Jason~D
  Lee}.} \bibinfo{year}{2018}\natexlab{}.
\newblock \showarticletitle{On the power of over-parametrization in neural
  networks with quadratic activation}.
\newblock \bibinfo{journal}{\emph{arXiv preprint arXiv:1803.01206}}
  (\bibinfo{year}{2018}).
\newblock


\bibitem[\protect\citeauthoryear{Eldan and Shamir}{Eldan and Shamir}{2016}]%
        {eldan2016power}
\bibfield{author}{\bibinfo{person}{Ronen Eldan} {and} \bibinfo{person}{Ohad
  Shamir}.} \bibinfo{year}{2016}\natexlab{}.
\newblock \showarticletitle{The power of depth for feedforward neural
  networks}. In \bibinfo{booktitle}{\emph{Conference on learning theory}}.
  \bibinfo{pages}{907--940}.
\newblock


\bibitem[\protect\citeauthoryear{Frey and Hinton}{Frey and Hinton}{1999}]%
        {frey1999variational}
\bibfield{author}{\bibinfo{person}{Brendan~J Frey} {and}
  \bibinfo{person}{Geoffrey~E Hinton}.} \bibinfo{year}{1999}\natexlab{}.
\newblock \showarticletitle{Variational learning in nonlinear Gaussian belief
  networks}.
\newblock \bibinfo{journal}{\emph{Neural Computation}} \bibinfo{volume}{11},
  \bibinfo{number}{1} (\bibinfo{year}{1999}), \bibinfo{pages}{193--213}.
\newblock


\bibitem[\protect\citeauthoryear{Gao and Ji}{Gao and Ji}{2019}]%
        {gao2019graph}
\bibfield{author}{\bibinfo{person}{Hongyang Gao} {and}
  \bibinfo{person}{Shuiwang Ji}.} \bibinfo{year}{2019}\natexlab{}.
\newblock \showarticletitle{Graph representation learning via hard and
  channel-wise attention networks}. In \bibinfo{booktitle}{\emph{Proceedings of
  the 25th ACM SIGKDD}}. \bibinfo{pages}{741--749}.
\newblock


\bibitem[\protect\citeauthoryear{Glorot and Bengio}{Glorot and Bengio}{2010}]%
        {glorot2010understanding}
\bibfield{author}{\bibinfo{person}{Xavier Glorot} {and} \bibinfo{person}{Yoshua
  Bengio}.} \bibinfo{year}{2010}\natexlab{}.
\newblock \showarticletitle{Understanding the difficulty of training deep
  feedforward neural networks}. In \bibinfo{booktitle}{\emph{Proceedings of the
  13th AISTATS}}. \bibinfo{pages}{249--256}.
\newblock


\bibitem[\protect\citeauthoryear{Goodfellow, Bengio, and Courville}{Goodfellow
  et~al\mbox{.}}{2016}]%
        {goodfellow2016deep}
\bibfield{author}{\bibinfo{person}{Ian Goodfellow}, \bibinfo{person}{Yoshua
  Bengio}, {and} \bibinfo{person}{Aaron Courville}.}
  \bibinfo{year}{2016}\natexlab{}.
\newblock \bibinfo{booktitle}{\emph{Deep learning}}.
\newblock \bibinfo{publisher}{MIT press}.
\newblock


\bibitem[\protect\citeauthoryear{Hawkins}{Hawkins}{2004}]%
        {hawkins2004problem}
\bibfield{author}{\bibinfo{person}{Douglas~M Hawkins}.}
  \bibinfo{year}{2004}\natexlab{}.
\newblock \showarticletitle{The problem of overfitting}.
\newblock \bibinfo{journal}{\emph{Journal of chemical information and computer
  sciences}} \bibinfo{volume}{44}, \bibinfo{number}{1} (\bibinfo{year}{2004}),
  \bibinfo{pages}{1--12}.
\newblock


\bibitem[\protect\citeauthoryear{Hayou, Doucet, and Rousseau}{Hayou
  et~al\mbox{.}}{2018}]%
        {hayou2018selection}
\bibfield{author}{\bibinfo{person}{Soufiane Hayou}, \bibinfo{person}{Arnaud
  Doucet}, {and} \bibinfo{person}{Judith Rousseau}.}
  \bibinfo{year}{2018}\natexlab{}.
\newblock \showarticletitle{On the selection of initialization and activation
  function for deep neural networks}.
\newblock \bibinfo{journal}{\emph{arXiv preprint arXiv:1805.08266}}
  (\bibinfo{year}{2018}).
\newblock


\bibitem[\protect\citeauthoryear{Hinton, Vinyals, and Dean}{Hinton
  et~al\mbox{.}}{2015}]%
        {hinton2015distilling}
\bibfield{author}{\bibinfo{person}{Geoffrey Hinton}, \bibinfo{person}{Oriol
  Vinyals}, {and} \bibinfo{person}{Jeff Dean}.}
  \bibinfo{year}{2015}\natexlab{}.
\newblock \showarticletitle{Distilling the Knowledge in a Neural Network}.
\newblock \bibinfo{journal}{\emph{stat}}  \bibinfo{volume}{1050}
  (\bibinfo{year}{2015}), \bibinfo{pages}{9}.
\newblock


\bibitem[\protect\citeauthoryear{Hornik, Stinchcombe, and White}{Hornik
  et~al\mbox{.}}{1989}]%
        {hornik1989multilayer}
\bibfield{author}{\bibinfo{person}{Kurt Hornik}, \bibinfo{person}{Maxwell
  Stinchcombe}, {and} \bibinfo{person}{Halbert White}.}
  \bibinfo{year}{1989}\natexlab{}.
\newblock \showarticletitle{Multilayer feedforward networks are universal
  approximators}.
\newblock \bibinfo{journal}{\emph{Neural networks}} \bibinfo{volume}{2},
  \bibinfo{number}{5} (\bibinfo{year}{1989}), \bibinfo{pages}{359--366}.
\newblock


\bibitem[\protect\citeauthoryear{Ioffe and Szegedy}{Ioffe and Szegedy}{2015}]%
        {ioffe2015batch}
\bibfield{author}{\bibinfo{person}{Sergey Ioffe} {and}
  \bibinfo{person}{Christian Szegedy}.} \bibinfo{year}{2015}\natexlab{}.
\newblock \showarticletitle{Batch normalization: Accelerating deep network
  training by reducing internal covariate shift}.
\newblock \bibinfo{journal}{\emph{arXiv preprint arXiv:1502.03167}}
  (\bibinfo{year}{2015}).
\newblock


\bibitem[\protect\citeauthoryear{Kalman and Kwasny}{Kalman and Kwasny}{1992}]%
        {kalman1992tanh}
\bibfield{author}{\bibinfo{person}{Barry~L Kalman} {and}
  \bibinfo{person}{Stan~C Kwasny}.} \bibinfo{year}{1992}\natexlab{}.
\newblock \showarticletitle{Why tanh: choosing a sigmoidal function}. In
  \bibinfo{booktitle}{\emph{[Proceedings 1992] IJCNN}},
  Vol.~\bibinfo{volume}{4}. IEEE, \bibinfo{pages}{578--581}.
\newblock


\bibitem[\protect\citeauthoryear{Kilian and Siegelmann}{Kilian and
  Siegelmann}{1993}]%
        {kilian1993power}
\bibfield{author}{\bibinfo{person}{Joe Kilian} {and} \bibinfo{person}{Hava~T
  Siegelmann}.} \bibinfo{year}{1993}\natexlab{}.
\newblock \showarticletitle{On the power of sigmoid neural networks}. In
  \bibinfo{booktitle}{\emph{Proceedings of the 6th annual conference on
  Computational learning theory}}. \bibinfo{pages}{137--143}.
\newblock


\bibitem[\protect\citeauthoryear{Krizhevsky and Hinton}{Krizhevsky and
  Hinton}{2009}]%
        {krizhevsky2009learning}
\bibfield{author}{\bibinfo{person}{Alex Krizhevsky} {and}
  \bibinfo{person}{Geoffrey Hinton}.} \bibinfo{year}{2009}\natexlab{}.
\newblock \showarticletitle{Learning multiple layers of features from tiny
  images}.
\newblock \bibinfo{journal}{\emph{Master's thesis, Department of Computer
  Science, University of Toronto}} (\bibinfo{year}{2009}).
\newblock


\bibitem[\protect\citeauthoryear{LeCun, Bottou, Bengio, et~al\mbox{.}}{LeCun
  et~al\mbox{.}}{1998}]%
        {lecun1998gradient}
\bibfield{author}{\bibinfo{person}{Yann LeCun}, \bibinfo{person}{L{\'e}on
  Bottou}, \bibinfo{person}{Yoshua Bengio}, {et~al\mbox{.}}}
  \bibinfo{year}{1998}\natexlab{}.
\newblock \showarticletitle{Gradient-based learning applied to document
  recognition}.
\newblock \bibinfo{journal}{\emph{Proc. IEEE}} \bibinfo{volume}{86},
  \bibinfo{number}{11} (\bibinfo{year}{1998}), \bibinfo{pages}{2278--2324}.
\newblock


\bibitem[\protect\citeauthoryear{Lu, Pu, Wang, Hu, and Wang}{Lu
  et~al\mbox{.}}{2017}]%
        {lu2017expressive}
\bibfield{author}{\bibinfo{person}{Zhou Lu}, \bibinfo{person}{Hongming Pu},
  \bibinfo{person}{Feicheng Wang}, \bibinfo{person}{Zhiqiang Hu}, {and}
  \bibinfo{person}{Liwei Wang}.} \bibinfo{year}{2017}\natexlab{}.
\newblock \showarticletitle{The expressive power of neural networks: A view
  from the width}. In \bibinfo{booktitle}{\emph{Advances in NIPS}}.
  \bibinfo{pages}{6231--6239}.
\newblock


\bibitem[\protect\citeauthoryear{Maass, Schnitger, and Sontag}{Maass
  et~al\mbox{.}}{1994}]%
        {maass1994comparison}
\bibfield{author}{\bibinfo{person}{Wolfgang Maass}, \bibinfo{person}{Georg
  Schnitger}, {and} \bibinfo{person}{Eduardo~D Sontag}.}
  \bibinfo{year}{1994}\natexlab{}.
\newblock \showarticletitle{A comparison of the computational power of sigmoid
  and Boolean threshold circuits}.
\newblock In \bibinfo{booktitle}{\emph{Theoretical Advances in Neural
  Computation and Learning}}. \bibinfo{publisher}{Springer},
  \bibinfo{pages}{127--151}.
\newblock


\bibitem[\protect\citeauthoryear{Montufar, Pascanu, Cho, and Bengio}{Montufar
  et~al\mbox{.}}{2014}]%
        {montufar2014number}
\bibfield{author}{\bibinfo{person}{Guido~F Montufar}, \bibinfo{person}{Razvan
  Pascanu}, \bibinfo{person}{Kyunghyun Cho}, {and} \bibinfo{person}{Yoshua
  Bengio}.} \bibinfo{year}{2014}\natexlab{}.
\newblock \showarticletitle{On the number of linear regions of deep neural
  networks}. In \bibinfo{booktitle}{\emph{Advances in NIPS}}.
  \bibinfo{pages}{2924--2932}.
\newblock


\bibitem[\protect\citeauthoryear{Nair and Hinton}{Nair and Hinton}{2010}]%
        {nair2010rectified}
\bibfield{author}{\bibinfo{person}{Vinod Nair} {and}
  \bibinfo{person}{Geoffrey~E Hinton}.} \bibinfo{year}{2010}\natexlab{}.
\newblock \showarticletitle{Rectified linear units improve restricted boltzmann
  machines}. In \bibinfo{booktitle}{\emph{Proceedings of the 27th ICML}}.
  \bibinfo{pages}{807--814}.
\newblock


\bibitem[\protect\citeauthoryear{Novak, Bahri, Abolafia, Pennington, and
  Sohl-Dickstein}{Novak et~al\mbox{.}}{2018}]%
        {novak2018sensitivity}
\bibfield{author}{\bibinfo{person}{Roman Novak}, \bibinfo{person}{Yasaman
  Bahri}, \bibinfo{person}{Daniel~A. Abolafia}, \bibinfo{person}{Jeffrey
  Pennington}, {and} \bibinfo{person}{Jascha Sohl-Dickstein}.}
  \bibinfo{year}{2018}\natexlab{}.
\newblock \showarticletitle{Sensitivity and Generalization in Neural Networks:
  an Empirical Study}. In \bibinfo{booktitle}{\emph{ICLR}}.
\newblock


\bibitem[\protect\citeauthoryear{Nwankpa, Ijomah, Gachagan, and
  Marshall}{Nwankpa et~al\mbox{.}}{2018}]%
        {nwankpa2018activation}
\bibfield{author}{\bibinfo{person}{Chigozie Nwankpa}, \bibinfo{person}{Winifred
  Ijomah}, \bibinfo{person}{Anthony Gachagan}, {and} \bibinfo{person}{Stephen
  Marshall}.} \bibinfo{year}{2018}\natexlab{}.
\newblock \showarticletitle{Activation functions: Comparison of trends in
  practice and research for deep learning}.
\newblock \bibinfo{journal}{\emph{arXiv preprint arXiv:1811.03378}}
  (\bibinfo{year}{2018}).
\newblock


\bibitem[\protect\citeauthoryear{Pascanu, Montufar, and Bengio}{Pascanu
  et~al\mbox{.}}{2013}]%
        {pascanu2013number}
\bibfield{author}{\bibinfo{person}{Razvan Pascanu}, \bibinfo{person}{Guido
  Montufar}, {and} \bibinfo{person}{Yoshua Bengio}.}
  \bibinfo{year}{2013}\natexlab{}.
\newblock \showarticletitle{On the number of response regions of deep feed
  forward networks with piece-wise linear activations}.
\newblock \bibinfo{journal}{\emph{arXiv preprint arXiv:1312.6098}}
  (\bibinfo{year}{2013}).
\newblock


\bibitem[\protect\citeauthoryear{Poole, Lahiri, Raghu, Sohl-Dickstein, and
  Ganguli}{Poole et~al\mbox{.}}{2016}]%
        {poole2016exponential}
\bibfield{author}{\bibinfo{person}{Ben Poole}, \bibinfo{person}{Subhaneil
  Lahiri}, \bibinfo{person}{Maithra Raghu}, \bibinfo{person}{Jascha
  Sohl-Dickstein}, {and} \bibinfo{person}{Surya Ganguli}.}
  \bibinfo{year}{2016}\natexlab{}.
\newblock \showarticletitle{Exponential expressivity in deep neural networks
  through transient chaos}. In \bibinfo{booktitle}{\emph{Advances in NIPS}}.
  \bibinfo{pages}{3360--3368}.
\newblock


\bibitem[\protect\citeauthoryear{Raghu, Poole, Kleinberg, Ganguli, and
  Dickstein}{Raghu et~al\mbox{.}}{2017}]%
        {raghu2017expressive}
\bibfield{author}{\bibinfo{person}{Maithra Raghu}, \bibinfo{person}{Ben Poole},
  \bibinfo{person}{Jon Kleinberg}, \bibinfo{person}{Surya Ganguli}, {and}
  \bibinfo{person}{Jascha~Sohl Dickstein}.} \bibinfo{year}{2017}\natexlab{}.
\newblock \showarticletitle{On the expressive power of deep neural networks}.
  In \bibinfo{booktitle}{\emph{Proceedings of the 34th ICML-Volume 70}}. JMLR,
  \bibinfo{pages}{2847--2854}.
\newblock


\bibitem[\protect\citeauthoryear{Silverman}{Silverman}{2018}]%
        {silverman2018density}
\bibfield{author}{\bibinfo{person}{Bernard~W Silverman}.}
  \bibinfo{year}{2018}\natexlab{}.
\newblock \bibinfo{booktitle}{\emph{Density estimation for statistics and data
  analysis}}.
\newblock \bibinfo{publisher}{Routledge}.
\newblock


\bibitem[\protect\citeauthoryear{Weng, Chung, and Szolovits}{Weng
  et~al\mbox{.}}{2019}]%
        {weng2019unsupervised}
\bibfield{author}{\bibinfo{person}{Wei-Hung Weng}, \bibinfo{person}{Yu-An
  Chung}, {and} \bibinfo{person}{Peter Szolovits}.}
  \bibinfo{year}{2019}\natexlab{}.
\newblock \showarticletitle{Unsupervised Clinical Language Translation}.
\newblock \bibinfo{journal}{\emph{Proceedings of the 25th ACM SIGKDD}}
  (\bibinfo{year}{2019}).
\newblock


\end{thebibliography}
